\documentclass{article}

\usepackage[margin=1in,footskip=0.25in]{geometry}

\usepackage[utf8]{inputenc} %
\usepackage[T1]{fontenc}    %
\usepackage{hyperref}       %
\usepackage{url}            %
\usepackage{booktabs}       %
\usepackage{amsfonts}       %
\usepackage{nicefrac}       %
\usepackage{microtype}      %
\usepackage{xcolor,colortbl}
\usepackage{tikz}
\usepackage[]{authblk}
\usepackage{graphicx, multirow, graphbox}
\usepackage{subfigure}
\usepackage{tabularx}
\usepackage{pifont}%
\graphicspath{ {figs/} }
\usepackage{comment}
\usepackage{amsmath,amssymb} %
\usepackage{color}
\usepackage{multirow}
\usepackage{needspace}
\usepackage{enumitem,kantlipsum}
\usepackage{wrapfig,lipsum}
\usepackage[numbers]{natbib}
\usepackage{makecell}
\usepackage{authblk}

\newif\ifarxiv
\arxivtrue

\ifarxiv 
\newcommand{\apdx}{\textit{Appendix}} 
\else \newcommand{\apdx}{\textit{Supp.~Material}} 
\fi

\title{\textbf{Graph Density-Aware Losses for Novel Compositions in Scene Graph Generation}}

\renewcommand\Affilfont{\small}

\author[1,2]{Boris Knyazev\footnote{This work was done while the author was an intern at Mila. Correspondence to: \texttt{bknyazev@uoguelph.ca}.}}
\author[3]{Harm de Vries}
\author[4]{C\u{a}t\u{a}lina Cangea}
\author[1,2,6]{\\Graham W.~Taylor}
\author[5,6,7]{Aaron Courville}
\author[5]{Eugene Belilovsky}
\affil[1]{School of Engineering, University of Guelph}
\affil[2]{Vector Institute for Artificial Intelligence}
\affil[3]{Element AI}
\affil[4]{University of Cambridge}
\affil[5]{Mila, Université de Montréal}
\makeatletter
\renewcommand\AB@affilsepx{, \protect\Affilfont}
\makeatother
\affil[6]{Canada CIFAR AI Chair}
\affil[7]{CIFAR LMB Fellow}

\newcommand{\eg}{\emph{e.g.}}

\newcommand{\std}[1]{{\tiny{$\pm$#1}}}
\newcommand\Tstrut{\rule{0pt}{3ex}}         %
\newcommand\Bstrut{\rule[-1.3ex]{0pt}{0pt}}   %

\begin{document}
	
	\date{\vspace{-5ex}}
	\maketitle

	\begin{abstract}
		
		Scene graph generation (SGG) aims to predict graph-structured descriptions of input images, in the form of objects and relationships between them. This task is becoming increasingly useful for progress at the interface of vision and language. Here, it is important---yet challenging---to perform well on novel (zero-shot) or rare (few-shot) compositions of objects and relationships. In this paper, we identify two key issues that limit such generalization. Firstly, we show that the standard loss used in this task is unintentionally a function of scene graph density. This leads to the neglect of individual edges in large sparse graphs during training, even though these contain diverse few-shot examples that are important for generalization. Secondly, the frequency of relationships can create a strong bias in this task, such that a ``blind'' model predicting the most frequent relationship achieves good performance. Consequently, some state-of-the-art models exploit this bias to improve results. We show that such models can suffer the most in their ability to generalize to rare compositions, evaluating two different models on the Visual Genome dataset and its more recent, improved version, GQA. To address these issues, we introduce a density-normalized edge loss, which provides more than a two-fold improvement in certain generalization metrics. Compared to other works in this direction, our enhancements require only a few lines of code and no added computational cost. We also highlight the difficulty of accurately evaluating models using existing metrics, especially on zero/few shots, and introduce a novel weighted metric.\footnote{The code is available at \url{https://github.com/bknyaz/sgg}.}
		
	\end{abstract}

	\section{Introduction}
	
\begin{figure}[t]
		\centering
		\begin{scriptsize}
			\setlength{\tabcolsep}{1pt}
			\begin{tabular}{c} \includegraphics[width=0.99\textwidth,align=c,trim={0 0.2cm 0 0.2cm},clip]{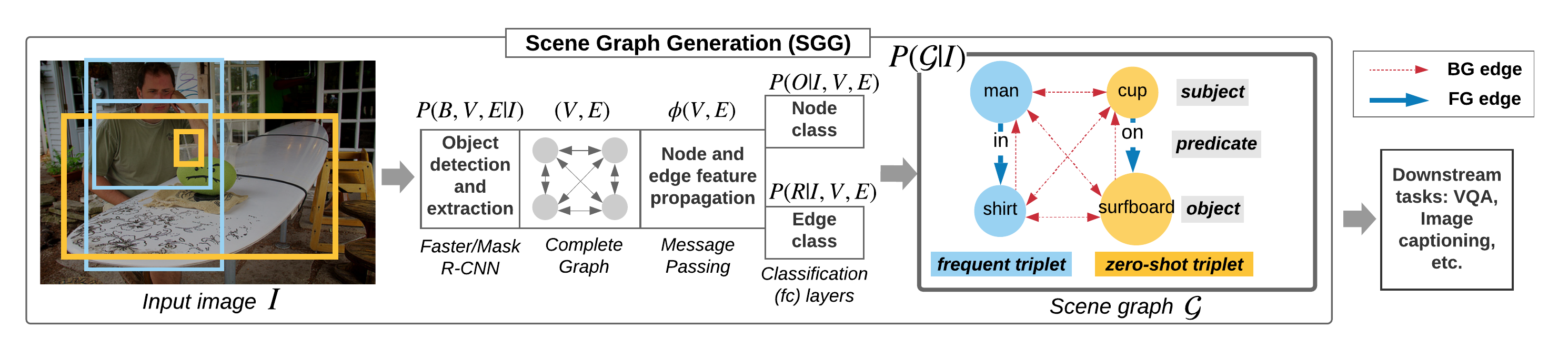} \\
			\end{tabular}
		\end{scriptsize}
		\vspace{-5pt}
		\caption{\small In this work, we improve scene graph generation $P({\cal G} | I)$. In many downstream tasks, such as VQA, the result directly depends on the accuracy of predicted scene graphs.}
		\label{fig:overview}
	\end{figure}

	In recent years, there has been growing interest to connect successes in visual perception with language and reasoning~\cite{su2019vl,zhou2019unified}. This requires us to design systems that can not only recognize objects, but understand and reason about the relationships between them. 
	This is essential for such tasks as visual question answering (VQA)~\cite{antol2015vqa,hudson2019gqa,cangea2019videonavqa} or caption generation~\cite{yang2019auto, gu2019unpaired}. 
	However, predicting a high-level semantic output (\eg~answer) from a low-level visual signal (\eg~image) is challenging due to a vast gap between the modalities.
	To bridge this gap, it would be useful to have some intermediate representation that can be relatively easily generated by the low-level module and, at the same time, can be effectively used by the high-level reasoning module.
	We want this representation to semantically describe the visual scene in terms of objects and relationships between them, which leads us to a structured image representation, the \textbf{scene graph} (SG)~\cite{johnson2015image, Krishna_2017}.
	A scene graph is a collection of visual relationship \textit{triplets}: <\textit{subject}, \textit{predicate}, \textit{object}> (\eg~<cup, on, table>). Each node in the graph corresponds to a subject or object (with a specific image location) and edges to predicates~(Figure~\ref{fig:overview}). 
	Besides bridging the gap, SGs can be used to verify how well the model has understood the visual world, as opposed to just exploiting one of the biases in a dataset~\cite{jabri2016revisiting,anand2018blindfold,bahdanau2018systematic}.
	Alternative directions to SGs include, for example, attention~\cite{norcliffe2018learning} and neural-symbolic models~\cite{vedantam2019probabilistic}.\looseness-1
	
	\begin{figure}
		\scriptsize
		\centering
		\setlength{\tabcolsep}{3pt}
		{
			\begin{minipage}{0.5\linewidth}
				\hspace{-20pt}
				\begin{tabular}{cc}
					\multicolumn{2}{c}{{\hspace{30pt}\includegraphics[width=0.95\textwidth,trim={3.5cm 8.3cm 7cm 0.8cm},clip]{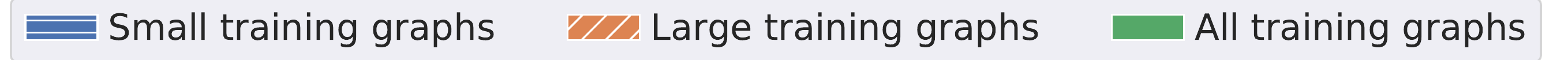}}} \vspace{-3pt} \\
					\vspace{-3pt}
					\includegraphics[align=c,height=2.7cm]{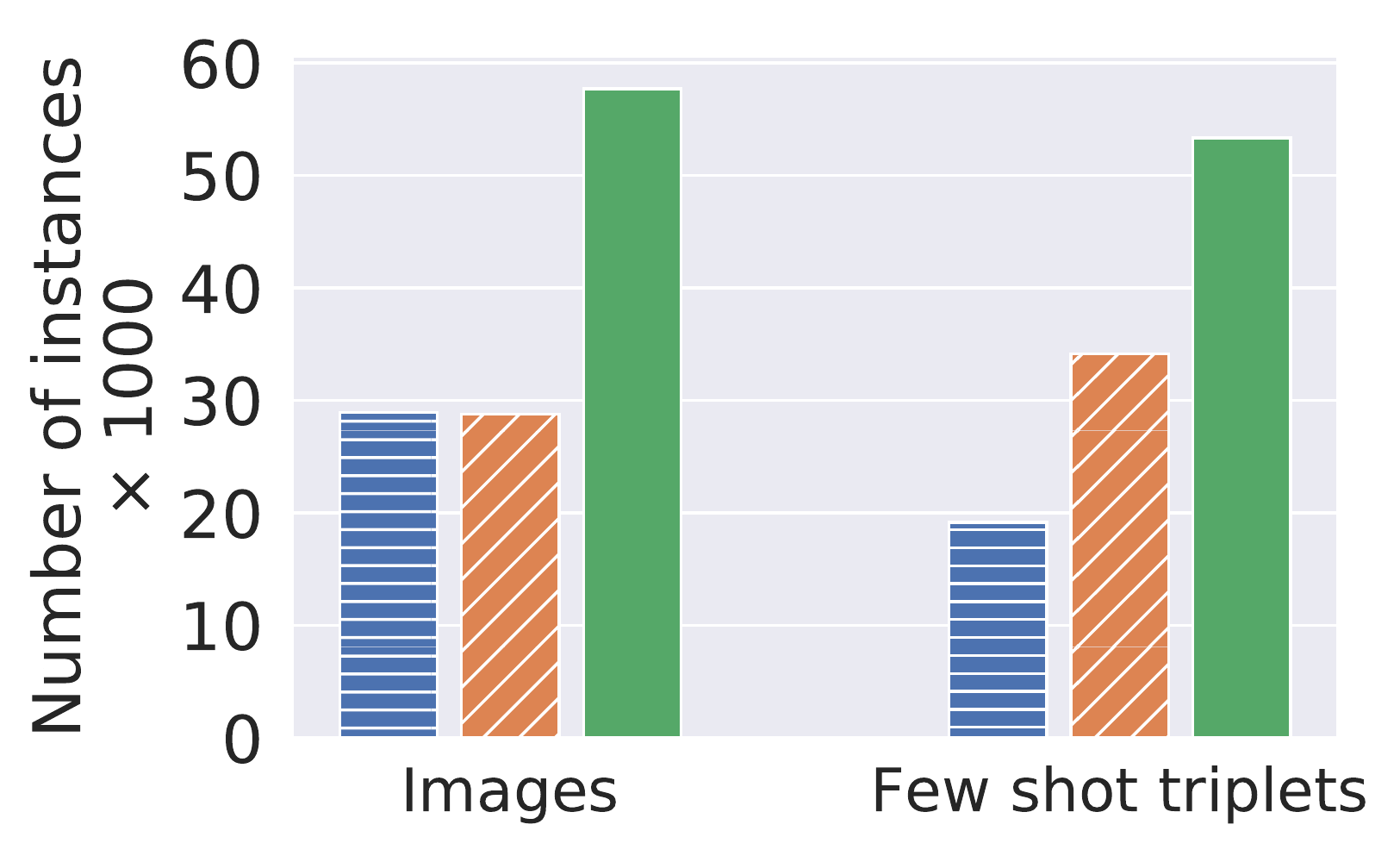} &
					\includegraphics[align=c,height=2.7cm]{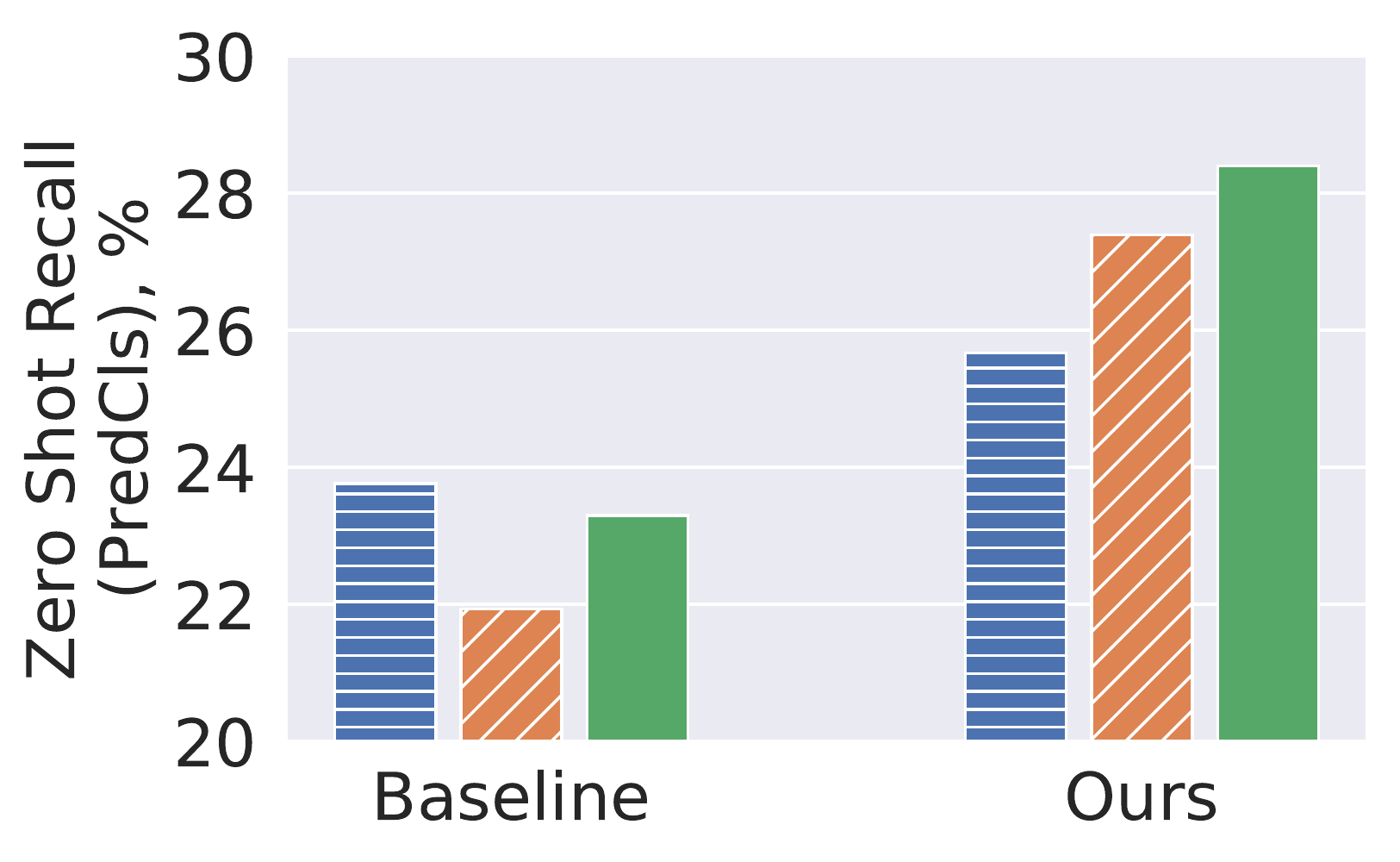} \\
					\hspace{20pt} (a) & \hspace{20pt} (b) 
				\end{tabular}
			\end{minipage}
		}
		\hspace{10pt}
		{
			\begin{minipage}{0.4\linewidth}
				\vspace{8pt}
				\begin{tabular}{p{3cm}p{3cm}}
					\centering
					\includegraphics[height=2cm, align=c]{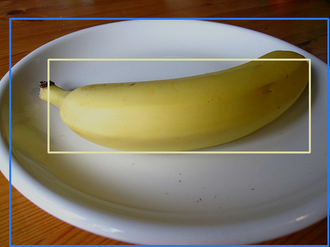} & \includegraphics[height=2cm, align=c]{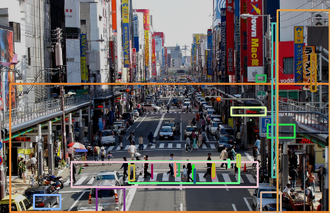} \Bstrut \\ 
					\centering (c) Image with a small scene graph & \centering (d) Image with a large scene graph
				\end{tabular}
			\end{minipage}
		}
		\vspace{-3pt}
		\caption{\small \textbf{Motivation of our work.} We split the training set of Visual Genome~\cite{Krishna_2017} into two subsets: those with relatively small ($\leq 10$ nodes) and large ($>10$ nodes) graphs. (\textbf{a}) While each subset contains a similar number of images (the three left bars), larger graphs contain more few-shot labels (the three right bars). (\textbf{b}) Baseline methods (\cite{xu2017scene} in this case) fail to learn from larger graphs due to their loss function. However, training on large graphs and corresponding few-shot labels is important for stronger generalization. We address this limitation and significantly improve results on zero and few-shots. (\textbf{c, d}) Small and large scene graphs typically describe simple and complex scenes respectively. \looseness-1 
		}
		\vspace{-10pt}
		\label{fig:motivation}
	\end{figure}

	\textbf{Scene graph generation} (SGG) is the task of predicting a SG given an input image. The inferred SG can be used directly for downstream tasks such as VQA~\cite{zhang2019empirical,NSM2019}, image captioning~\cite{yang2019auto, gu2019unpaired} or retrieval~\cite{johnson2015image,belilovsky2017joint,tang2020unbiased}.
	A model which performs well on SGG should demonstrate the ability to ground visual concepts to images and generalize to compositions of objects and predicates in new contexts. In real world images, some compositions (\eg~<cup, on, table> ) appear more frequently than others (\eg~<cup, on, \textit{surfboard}> or <cup, \textit{under}, table>), which creates a strong frequency bias. This makes it particularly challenging for models to generalize to novel (zero-shot) and rare (few-shot) compositions, even though each of the subjects, objects and predicates have been observed at training time.
	The problem is exacerbated by the test set and evaluation metrics, which do not penalize models that blindly rely on such bias. Indeed,~\citet{zellers2018neural} has pointed out that SGG models largely exploit simple co-occurrence information. In fact, the performance of models predicting solely based on frequency  (i.e. a cup is most likely to be \textit{on} a \textit{table}) is not far from the state-of-the-art using common metrics (see \textsc{Freq} in Table~\ref{table:main_results}).\looseness-1
	
	In this work, we reveal that (a) the frequency bias exploited by certain  models leads to poor generalization on few-shot and zero-shot compositions; (b) existing models disproportionately penalize large graphs, even if these often contain many of the infrequent visual relationships, which leads to performance degradation on few and zero-shot cases (Figure~\ref{fig:motivation}). We address these challenges and show that our suggested improvements can provide benefits for two strong baseline models~\cite{xu2017scene,zellers2018neural}. Overall, we make the following four \textbf{contributions}:
	
	\begin{enumerate}[leftmargin=10pt,labelsep=2pt]
		\itemsep0em 
		\item \textbf{Improved loss}: we introduce a density-normalized edge loss, which improves results on all metrics, especially for few and zero-shots (Section~\ref{sec:loss_norm});
		\item \textbf{Novel weighted metric}: we illustrate several issues in the evaluation of few and zero-shots, proposing a novel weighted metric which can better track the performance of this critical desiderata (Section~\ref{sec:metric});
		\item \textbf{Frequency bias}: we demonstrate a negative effect of the frequency bias, proposed in Neural Motifs~\cite{zellers2018neural}, on few and zero-shot performance (Section~\ref{sec:experiments}); 
		\item \textbf{Scaling to GQA}: in addition to evaluating on Visual Genome (VG)~\cite{Krishna_2017}, we confirm the usefulness of our loss and metrics on GQA~\cite{hudson2019gqa} -- an improved version of VG. GQA has not been used to evaluate SGG models before and is interesting to study, because compared to VG its scene graphs are cleaner, larger, more dense and contain a larger variety of objects and predicates (Section~\ref{sec:experiments}).
	\end{enumerate}
	\vspace{-10pt}

	\section{Related Work}
	
	\textbf{Zero-shot learning.}
	In vision tasks, such as image classification, zero-shot learning has been extensively studied, and the main approaches are based on attributes~\cite{lampert2013attribute} and semantic embeddings~\cite{frome2013devise, xian2016latent}.
	The first approach is related to the zero-shot problem we address in this work: it assumes that all individual attributes of objects (color, shape, etc.) are observed during training, such that novel classes can be detected at test time based on \textit{compositions} of their attributes.
	Zero-shot learning in scene graphs is similar: all individual subjects, objects and predicates are observed during training, but most of their compositions are not. This task was first evaluated in~\cite{lu2016visual} on the VRD dataset using a joint vision-language model. 
	Several follow-up works attempted to improve upon it: by learning a translation operator in the embedding space~\cite{zhang2017visual}, clustering in a weakly-supervised fashion~\cite{peyre2017weakly}, using conditional random fields~\cite{cong2018scene} or optimizing a cycle-consistency loss to learn object-agnostic features~\cite{yang2018shuffle}.
	Augmentation using generative networks to generate more examples of rare cases is another promising approach~\cite{wang2019generating}; but, it was only evaluated in the predicate classification task. In our work, we also consider subject/object classification to enable the classification of the whole triplets, making the ``image to scene graph'' pipeline complete. Most recently,~\citet{tang2020unbiased} proposed learning causal graphs and showed strong performance in zero-shot cases.
	
	While these works improve generalization, none of them has identified the challenges and importance of learning from large graphs for generalization. By concentrating the model's capacity on smaller graphs and neglecting larger graphs, baseline models limit the variability of training data, useful for stronger generalization~\cite{hill2019environmental}. Our loss enables this learning, increasing the effective data variability. Moreover, previous gains typically incur a large computational cost, while our loss has negligible cost and can be easily added to other models.\looseness-1

	\textbf{Few-shot predicates.} 
	Several recent works have addressed the problem of imbalanced and few-shot predicate classes ~\cite{chen2019knowledge, dornadula2019visual,tang2019learning,zhang2019graphical,tang2020unbiased,chen2019scene}. However, compared to our work, these works have not considered the imbalance between foreground and background edges, which is more severe than other predicate classes (Figure~\ref{fig:pred_distr}) and is important to be fixed as we show in this work.
	Moreover, we argue that the compositional generalization, not addressed in those works, can be more difficult than generalization to rare predicates. For example, the triplet <cup, on, surfboard> is challenging to be predicted correctly as a whole; even though `on' can be the most frequent predicate, it has never been observed together with `cup' and `surfboard'. Experimental results in previous work~\cite{lu2016visual,zhang2017visual,yang2018shuffle,wang2019generating,tang2020unbiased} highlight this difficulty. Throughout this work, by ``few-shot'' we assume triplets, not predicates.
	
	\textbf{``Unbiasing'' methods.}
	Our idea is similar to the Focal loss~\cite{lin2017focal}, which addresses the imbalance between foreground and background objects in the object detection task. However, directly applying the focal loss to Visual Genome is challenging, due to the large amount of missing and mislabeled examples in the dataset. In this case, concentrating the model's capacity on ``hard'' examples can be equivalent to putting more weight on noise, which can hurt performance. \citet{tang2020unbiased} compared the focal loss and other unbiasing methods, such as upsampling and upweighting, and did not report significantly better results.

	\section{Methods}
	\label{sec:methods}
	In this section, we will review a standard loss used to train scene graph generation models (Section~\ref{sec:baseline}) and describe our improved loss (Section~\ref{sec:loss_norm}). We will then discuss issues with evaluating rarer combinations and propose a new weighted metric (Section~\ref{sec:metric}).

	\subsection{Overview of Scene Graph Generation}
	\label{sec:baseline}
	In scene graph generation, given an image $I$, we aim to output a scene graph ${\cal G}=(O,R)$ consisting of a set of subjects and objects ($O$) as nodes and a set of relationships or predicates ($R$) between them as edges (Figure~\ref{fig:overview}). So the task is to maximize the probability $P({\cal G} | I)$, which can be expressed as  $P(O,R | I) = P(O |I)\ P(R | I,O)$.
	Except for works that directly learn from pixels~\cite{newell2017pixels}, the task is commonly~\cite{xu2017scene,yang2018graph,zellers2018neural} reformulated by first detecting bounding boxes $B$ and extracting corresponding object and edge features, $V=f(I,B)$ and $E=g(I,B)$ respectively, using some functions $f,g$ (\eg~a ConvNet followed by ROI Align~\cite{he2017mask}):
	\begin{equation}
	\label{eq:scene_graph_prob_box}
	P({\cal G} |I) = P(V,E |I)\ P(O,R | V,E,I).%
	\end{equation}
	The advantage of this approach is that solving $P(O,R | V,E,I)$ is easier than solving $P(O,R | I)$. At the same time, to compute $P(V,E |I)$ we can use pretrained object detectors~\cite{ren2015faster, he2017mask}.
	Therefore, we follow~\cite{xu2017scene,yang2018graph,zellers2018neural} and use this approach to scene graph generation.
	
	In practice, we can assume that the pretrained object detector is fixed or that ground truth bounding boxes $B$ are available, so we can assume $P(V,E|I)$ is constant. In addition, following~\cite{lu2016visual,xu2017scene,yang2018graph}, we can assume conditional independence of variables $O$ and $R$: $P(O,R|V,E,I)=P(O|V,E,I)P(R|V,E,I)$. We thus obtain the scene graph generation loss:
	\begin{equation}
	\label{eq:scene_graph_prob_simple}
	-\text{log} P({\cal G} |I) = -\text{log} P(O | V,E,I) - \text{log} P(R | V,E,I).
	\end{equation}
	Some models~\cite{zellers2018neural,zhang2019graphical} do not assume the conditional independence of $O$ and $R$, making predicates explicitly depend on subject and object labels: $P(O,R|V,E,I)=P(O|V,E,I) P(R|O,V,E,I)$. However, such a model must be carefully regularized, since it can start to ignore $(V,E,I)$ and mainly rely on the frequency distribution $P(R|O)$ as a stronger signal. For example, the model can learn that between `cup' and `table' the relationship is most likely to be `on', regardless the visual signal. As we show, this can hurt generalization.\looseness-1
	
	\begin{wrapfigure}{r}{5.5cm}%
		\vspace{-5pt}
		\centering
		\includegraphics[width=5.5cm,align=c]{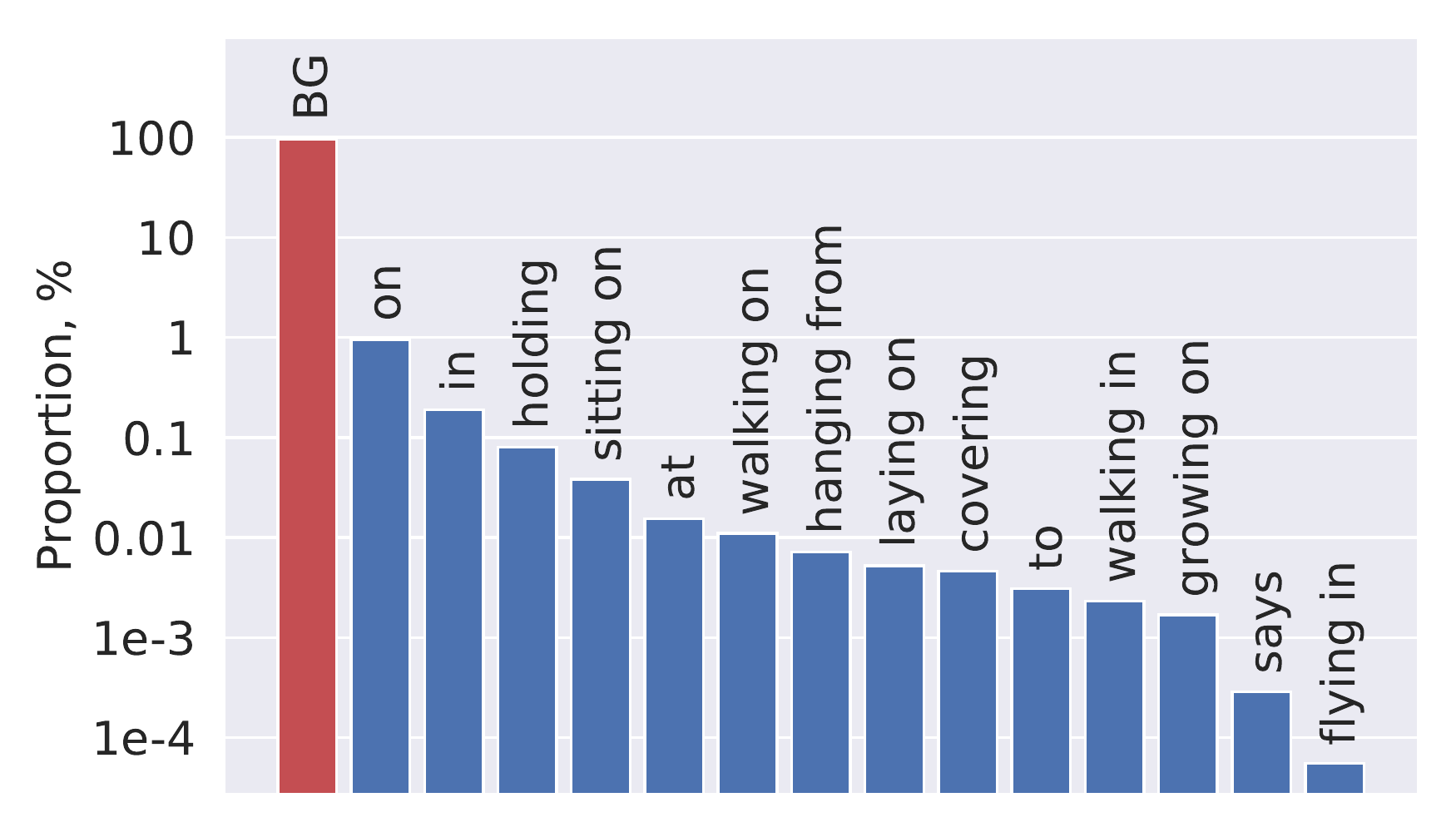} \\
		\caption{\small Predicate distribution in Visual Genome (split~\cite{xu2017scene}). BG edges (note the log scale) dominate, with >96\% of all edges, creating an extreme imbalance. For clarity, only some most and least frequent predicate classes are shown.\looseness-1}
		\label{fig:pred_distr}
	\end{wrapfigure}
	
	Eq.~\eqref{eq:scene_graph_prob_simple} is commonly handled as a multitask classification problem, where each task is optimized by the cross-entropy loss ${\cal L}$.
	In particular, given a batch of scene graphs with $N$ nodes and $M$ edges in total, the loss is the following:
	\begin{equation}
	\label{eq:loss_baseline}
	{\cal L} = {\cal L}_{node} + {\cal L}_{edge} = \frac{1}{N}\sum_i^N{\cal L}_{obj,i} + \frac{1}{M}\sum_{ij}^M {\cal L}_{rel,ij}.
	\end{equation}
	Node and edge features $(V,E)$ output by the detector form a complete graph without self-loops (Figure~\ref{fig:overview}). So, conventionally~\cite{xu2017scene,yang2018graph,zellers2018neural}, the loss is applied to all edges: $M \approx N^2$.
	These edges can be divided into foreground (FG), corresponding to annotated edges, and background (BG), corresponding to not annotated edges: $M = M_{FG} + M_{BG}$.
	The BG edge type is similar to a ``negative'' class in the object detection task and has a similar purpose.
	Without training on BG edges, at test time the model would label all pairs of nodes as ``positive'', i.e. having some relationship, when often it is not the case (at least, given the vocabulary in the datasets). Therefore, not using the BG type can hurt the quality of predicted scene graphs and can lower recall.\looseness-1
	
	\subsection{Hyperparameter-free Normalization of the Edge Loss}
	\label{sec:loss_norm}
	
\begin{figure}[h]
		\centering
		\begin{small}
			\setlength{\tabcolsep}{2pt}
			\begin{tabular}{ccc}
				\includegraphics[height=3.5cm,align=c]{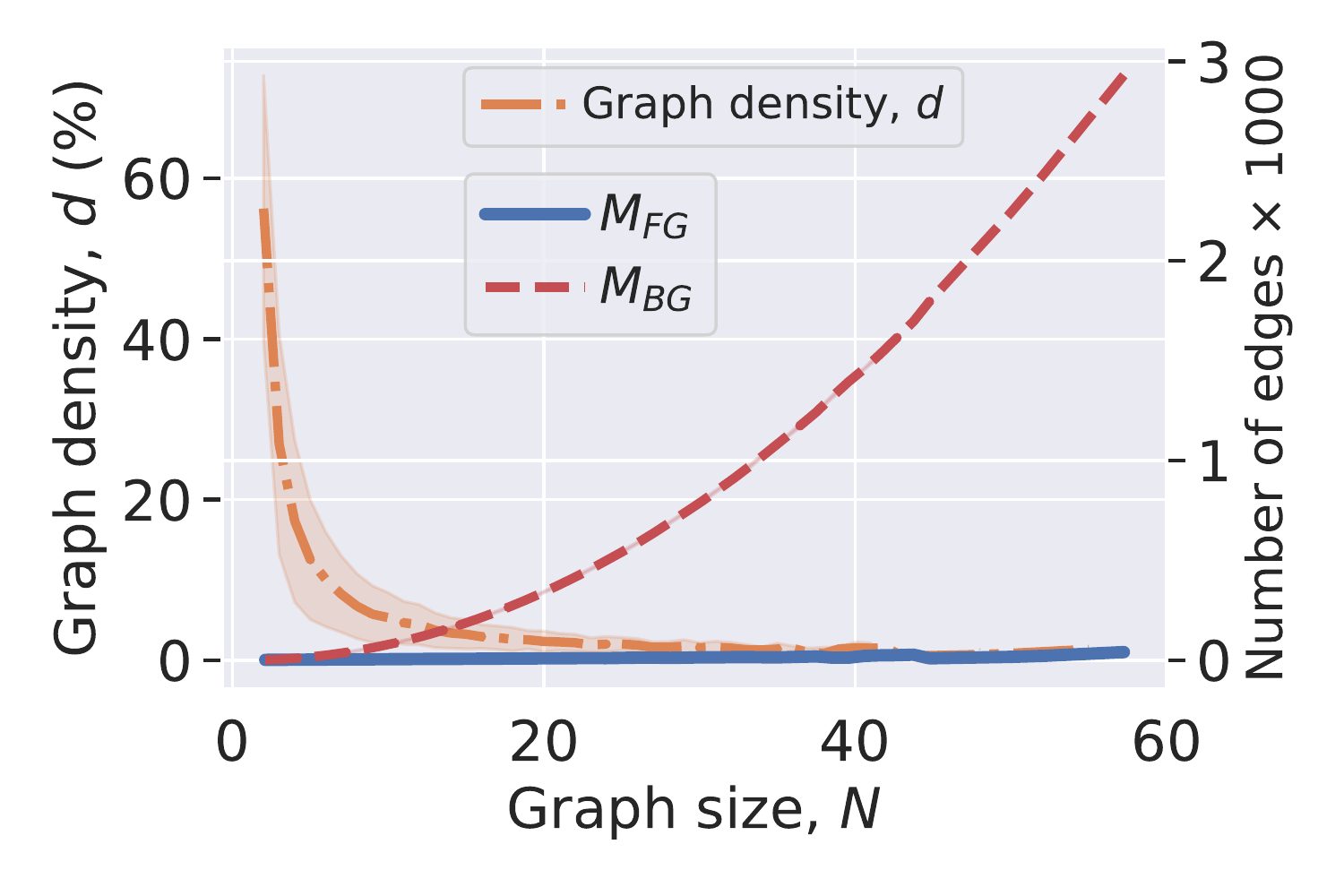} &
				\includegraphics[height=3.5cm,align=c]{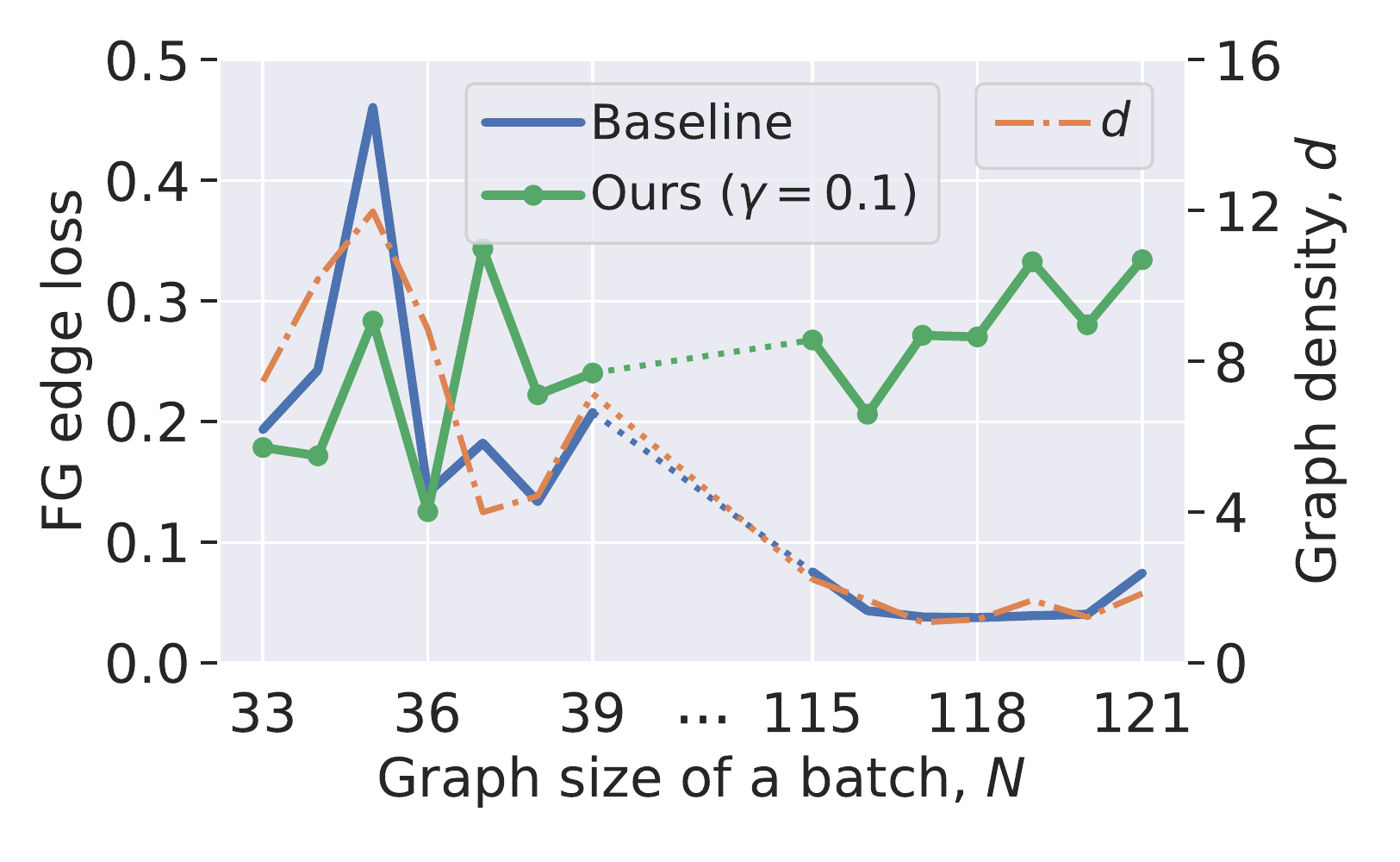} & \includegraphics[height=3.5cm,align=c]{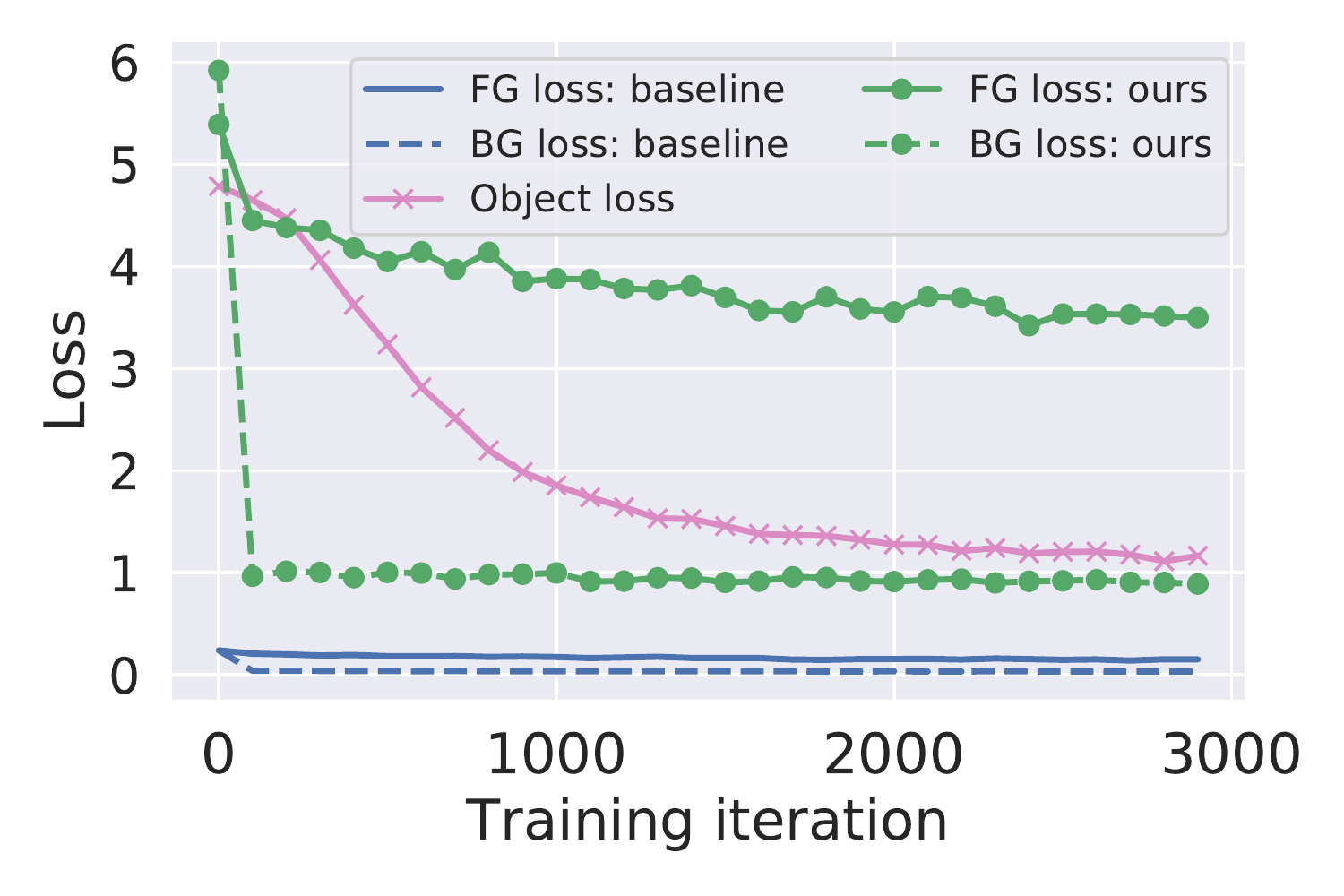} \\ $d \rightarrow 0 $ for large graphs & Large graphs are downweighted by $d$ & $L_{edge} \ll L_{node}$ \\
			\end{tabular}
		\end{small}
		\vspace{-5pt}
		\caption{
			(\small \textbf{left}) The number of FG edges grows much more slowly with graph size than the number of BG edges (on VG: $M_{FG}\approx 0.5N$, see plots for different batch sizes and the GQA dataset in Figure~\ref{fig:graph_density_more} in~\apdx). This leads to: (\textbf{middle}) Downweighting of the FG loss on larger graphs and effectively limiting the amount and variability of training data, since large graphs contain a lot of labeled data (Figure~\ref{fig:motivation}); here, losses of converged models for batches sorted by graph size are shown. (\textbf{right}) Downweighting of the edge loss $L_{edge}$ overall compared to $L_{node}$, even though both tasks are equally important to correctly predicting a scene graph. Our normalization fixes both issues.}
		\label{fig:graph_density}
	\end{figure}

	\textbf{Baseline loss as a function of graph density.}
	In scene graph datasets such as Visual Genome, the number of BG edges is greater than FG ones (Figure~\ref{fig:pred_distr}), yet the baseline loss~\eqref{eq:loss_baseline} does not explicitly differentiate between BG and other edges. If we assume a fixed probability for two objects to have a relationship, then as the number of nodes grows we can expect fewer of them to have a relationship. Thus the graph density can vary based on the number of nodes (Figure~\ref{fig:graph_density}), a fact not taken into account in Eq.~\eqref{eq:loss_baseline}.
	To avoid this, we start by decoupling the edge term of~\eqref{eq:loss_baseline} into the foreground (FG) and background (BG) terms:
	\begin{equation}
	{\cal L}_{edge} = \frac{1}{M}\sum_{ij}^M{\cal L}_{rel,ij} =  \frac{1}{M_{FG} + M_{BG}} \bigg[ \underbrace{\sum\nolimits_{ij \in \cal E}^{M_{FG}}{\cal L}_{rel,ij}}_{\text{FG edges}} + \underbrace{\sum\nolimits_{ij \notin \cal E}^{M_{BG}}{\cal L}_{rel,ij}}_{\text{BG edges}} \bigg]  ,
	\end{equation}
	\noindent where ${\cal E}$ is a set of FG edges, $M_{FG}$ is the number of FG edges ($|{\cal E}|$) and $M_{BG}$ is the number of BG edges.
	Next, we denote FG and BG edge losses averaged per batch as ${\cal L}_{FG}=1/M_{FG}\sum_{ij \in \cal E}^{M_{FG}}{\cal L}_{rel,ij}$ and ${\cal L}_{BG}=1/M_{BG}\sum_{ij \notin \cal E}^{M_{BG}}{\cal L}_{rel,ij}$, respectively.
	Then, using the definition of graph density as a proportion of FG edges to all edges, $d=M_{FG} / (M_{BG} + M_{FG})$, we can express the total baseline loss equivalent to Eq.~\eqref{eq:loss_baseline} \textbf{as a function of graph density}:
	\begin{equation}
	\label{eq:loss_decouple}
	\boxed{
		{\cal L} = {\cal L}_{node} + d {\cal L}_{FG}  + (1 - d) {\cal L}_{BG}.}
	\end{equation}

	\noindent\textbf{Density-normalized edge loss.}
	Eq.~\eqref{eq:loss_decouple} and Figure~\ref{fig:graph_density} allow us to notice two issues:
	
	\begin{enumerate}[leftmargin=10pt,labelsep=2pt]%
		\itemsep0em 
		\item \textbf{Discrepancy of the loss between graphs of different sizes.} Since $d$ exponentially decreases with graph size (Fig.~\ref{fig:graph_density}, left), FG edges of larger graphs are weighted less than edges of smaller graphs in the loss (Fig.~\ref{fig:graph_density}, middle), making the model neglect larger graphs.\looseness-1
		
		\item \textbf{Discrepancy between object and edge losses.}
		Due to $d$ tending to be small on average, $L_{edge}$ is much smaller than $L_{node}$, so the model might focus mainly on $L_{node}$ (Fig.~\ref{fig:graph_density}, right).
	\end{enumerate}
	
	\vspace{-5pt}
	Both issues can be addressed by normalizing FG and BG terms by graph density $d$:
	\begin{equation}
	\label{eq:loss_norm}
	\boxed{{\cal L} = {\cal L}_{node} + \gamma \big[ {\cal L}_{FG}  + {M_{BG}}/{M_{FG}} {\cal L}_{BG}\big].}
	\end{equation}
	\noindent where $\gamma=1$ in our default hyperparameter-free variant and $\gamma \neq 1$ only to empirically analyze the loss (Table~\ref{table:loss_ablations}). Even though the BG term still depends on graph density, we found it to be less sensitive to variations in $d$, since the BG loss quickly converges to some stable value, performing a role of regularization (Figure~\ref{fig:graph_density}, right).
	We examine this in detail in Section~\ref{sec:experiments}.
	
	\subsection{Weighted Triplet Recall}
	\label{sec:metric}
	
\begin{figure}[h]
		\centering
		\vspace{-10pt}
		{\includegraphics[width=\textwidth]{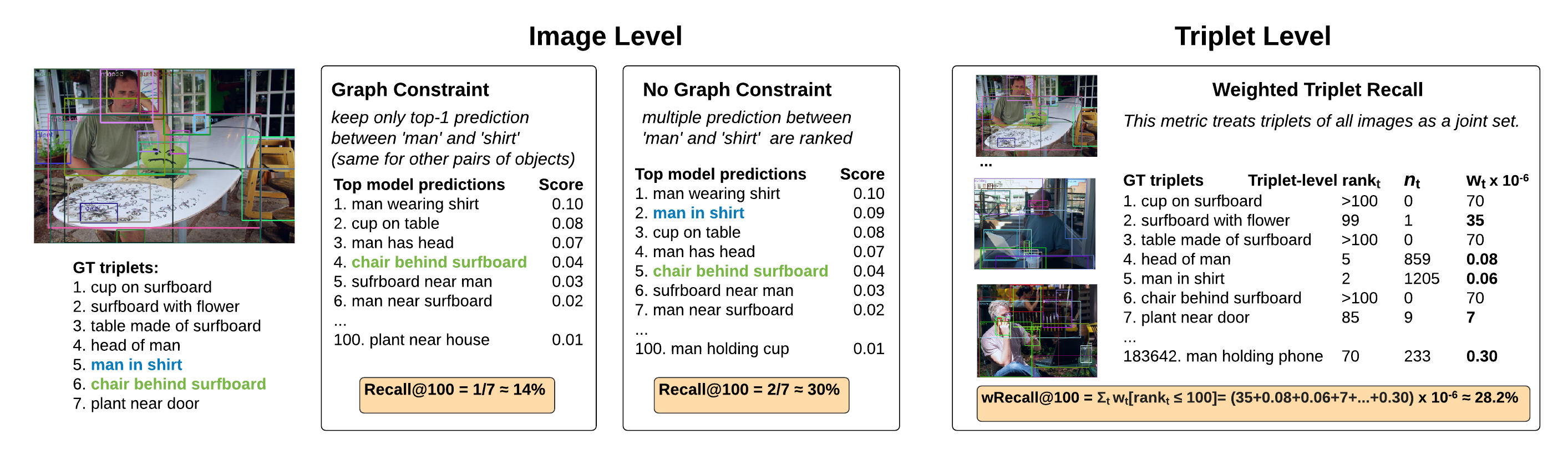}}
		\vspace{-25pt}
		\caption{\small Existing image-level recall metrics \textit{versus} our proposed weighted triplet recall. We first make unweighted predictions $\text{rank}_t \leq K$ for all GT triplets in all test images, then reweight them according to the frequency distribution~\eqref{eq:recall_triplet}. Computing our metric per image would be noisy.
		}
		\label{fig:eval}
	\end{figure}

	The common evaluation metric for scene graph prediction is image-level Recall@K or R@K \cite{xu2017scene, yang2018graph, zellers2018neural}.
	To compute it, we first need to extract the top-$K$ triplets, $\text{Top}_K$, from the \textit{entire image} based on ranked predictions of a model . Given a set of ground truth triplets, \text{GT}, the image-level R@K is computed as (see Figure~\ref{fig:eval} for a visualization): 
	\setlength{\abovedisplayskip}{3pt}
	\setlength{\belowdisplayskip}{3pt}
	\begin{equation}
	\label{eq:recall_image}
	\text{R}@\text{K} = {|\text{Top}_K \cap \text{GT}|}/{|\text{GT}|}.
	\end{equation}
	There are four issues with this metric (we discuss additional details in~\apdx):
	
	\begin{enumerate}[label=(\alph*),wide, labelwidth=!, labelindent=0pt, leftmargin=0pt, listparindent=0pt, labelsep=3pt]
		\itemsep0em 
		\item The frequency bias of triplets means more frequent triplets will dominate the metric.
		\item The denominator in~\eqref{eq:recall_image} creates discrepancies between images with different $|\text{GT}|$ (the number of ground truth triplets in an image), especially pronounced in few/zero shots.
		\item Evaluation of zero ($n=0$) and different few-shot cases $n=1,5,10,...$~\cite{wang2019generating} leads to many R@K results~\cite{wang2019generating}. This complicates the analysis. Instead, we want a single metric for all $n$.
		\item Two ways of computing the image-level recall~\cite{newell2017pixels,zellers2018neural}, graph \textit{constrained} and \textit{unconstrained}, lead to very different results and complicate the comparison (Figure~\ref{fig:eval}).%
	\end{enumerate}
	
	To address issue \textbf{(a)}, the predicate-normalized metric, mean recall (mR@K)~\cite{chen2019knowledge,tang2019learning} and weighted mR@K were introduced~\cite{zhang2019graphical}. These metrics, however, only address the imbalance of predicate classes, not whole triplets.
	Early work~\cite{lu2016visual,dai2017detecting} used triplet-level Recall@K (or R$_{tr}$@K) for some tasks (\eg~predicate detection), which is based on ranking predicted triplets for each ground truth subject-object pair independently; the pairs without relationships are not evaluated. Hence, R$_{tr}$@K is similar to top-$K$ accuracy.
	This metric avoids issues \textbf{(b)} and \textbf{(d)}, but the issues of the frequency bias \textbf{(a)} and unseen/rare cases \textbf{(c)} still remain.
	To alleviate these, we adapt this metric to better track unseen and rare cases. We call our novel metric Weighted Triplet Recall wR$_{tr}$@K, which computes a recall at each triplet and reweights the average result based on the frequency of the GT triplet in the training set: 
	\setlength{\abovedisplayskip}{3pt}
	\setlength{\belowdisplayskip}{3pt}
	\begin{equation}
	\label{eq:recall_triplet}
	\text{wR}_{tr}@\text{K} =  \sum\nolimits_t^T w_t [\text{rank}_t \leq K],
	\end{equation}
	\noindent where $T$ is the number of all test triplets, $[\cdot]$ is the Iverson bracket, $w_t=\frac{1}{(n_t + 1)\sum_t 1/(n_t + 1)} \in [0,1]$ and $n_t$ is the number of occurrences of triplet $t$ in the training set; $n_t + 1$ is used to handle zero-shot triplets; $\sum_t w_t = 1$. Since wR$_{tr}$@K is still a triplet-level metric, we avoid issues \textbf{(b)} and \textbf{(d)}. Our metric is also robust to the frequency-bias \textbf{(a)}, since frequent triplets (with high $n_t$) are downweighted proportionally, which we confirm by evaluating the \textsc{Freq} model from~\cite{zellers2018neural}. Finally, a single wR$_{tr}$@K value shows zero and few-shot performance linearly aggregated for all $n \geq 0$, solving issue \textbf{(c)}. 
	
	\vspace{-5pt}
	\section{Experiments}
	\label{sec:experiments}
	\vspace{-5pt}
	
	\textbf{Datasets.}
	We evaluate our loss and metric on Visual Genome~\cite{Krishna_2017}. Since it is a noisy dataset, several ``clean'' variants were introduced. We mainly experiment with the most common variant (VG)~\cite{xu2017scene}, which consists of the 150 most frequent object and 50 predicates classes. An alternative variant (VTE)~\cite{zhang2017visual} has been often used for zero-shot evaluation. Surprisingly, we found that the VG split~\cite{xu2017scene} is better suited for this task, given a larger variability of zero-shot triplets in the test set (see Table~\ref{table:datasets} in~\apdx). Recently, GQA~\cite{hudson2019gqa} was introduced, where scene graphs were cleaned to automatically construct question answer pairs. GQA has more object and predicate classes, so that zero and few-shot triplets are more likely to occur at test time. To the best of our knowledge, scene graph generation (SGG) results have not been reported on GQA before, even though some VQA models have relied on SGG~\cite{NSM2019}.
	
	\textbf{Training and evaluation details.}
	We experiment with two models: Message Passing (MP)~\cite{xu2017scene} and Neural Motifs (NM)~\cite{zellers2018neural}. We use publicly available implementations of MP and NM\footnote{\url{https://github.com/rowanz/neural-motifs}}, with all architecture details and hyperparameters kept the same (except for the small changes outlined in Table~\ref{table:arch} in~\apdx).
	To be consistent with baseline models, for Visual Genome we use Faster R-CNN~\cite{ren2015faster} with VGG16 as a backbone to extract node and edge features. For GQA we choose a more recent Mask R-CNN~\cite{he2017mask} with ResNet-50-FPN as a backbone pretrained on COCO. We also use this detector on the VTE split.
	We perform more experiments with Message Passing, since our experiments revealed that it better generalizes to zero and few-shot cases, while performing only slightly worse on other metrics. In addition, it is a relatively simple model, which makes the analysis of its performance easier.
	We evaluate models on three tasks, according to~\cite{xu2017scene}: 1) predicate classification (\textbf{PredCls}), in which the model only needs to label a predicate given ground truth object labels and bounding boxes, i.e. $P(R|I,B,O)$; 2) scene graph classification (\textbf{SGCls}), in which the model must also label objects, i.e. $P(O,R|I,B)$; 3) scene graph generation \textbf{SGGen} (sometimes denoted as \textbf{SGDet}), $P({\cal G}|I)$, which includes detecting bounding boxes first (reported separately in Tables~\ref{table:sggen_results},~\ref{table:gqa_sgg}).
	
	\subsection{Results}
	Table~\ref{table:main_results} shows our main results, where for each task we report five metrics: image-level recall on all triplets (R@K) and zero-shot triplets (R$_{ZS}$@K), triplet-level recall (R$_{tr}$@K) and our weighted triplet recall (wR$_{tr}$@K), and mean recall (mR@K). We compute recalls without the graph constraint since, as we discuss in~\apdx, this is a more accurate metric. We denote graph-constrained results as \textbf{PredCls-GC}, \textbf{SGCls-GC}, \textbf{SGGen-GC} and report them only in Tables~\ref{table:zs_comparison_vte},~\ref{table:zs_comparison_vg} and Table~\ref{table:sggen_results}.
	
	\definecolor{extreme}{rgb}{0.82,0.82,0.90}
	\definecolor{bad}{rgb}{0.9,0.9,0.9}
	\newcommand\crule[3][black]{\textcolor{#1}{\rule{#2}{#3}}}
	\newcommand{\rel}[1]{{\tiny{+#1\%}}}
	
	\setlength{\tabcolsep}{3pt}
	\begin{table}[t]
		\vspace{-10pt}
		\scriptsize
		\begin{center}
			\begin{tabular}{p{0.4cm}|llcccccp{0.1cm}ccccc}
				\multirow{2}{*}{\rotatebox[origin=c]{90}{\hspace{1pt}\centering\tiny\parbox{0.5cm}{\vspace{0pt}\textbf{Data\\-set}}}} & \multirow{2}{*}{\textbf{Model}} & \multirow{2}{*}{\textbf{Loss}} & \multicolumn{5}{c}{\textbf{Scene Graph Classification}} & & \multicolumn{5}{c}{\textbf{Predicate Classification}} \\
				\cline{4-8}\cline{10-14}
				& & & \tiny R@100 & \tiny R$_{ZS}$@100 & \tiny R$_{tr}$@20& \tiny wR$_{tr}$@20 & \tiny mR@100 &
				& \tiny R@50 & \tiny R$_{ZS}$@50 & \tiny R$_{tr}$@5 & \tiny wR$_{tr}$@5 & \tiny mR@50 \Tstrut \Bstrut\\
				\Xhline{2\arrayrulewidth} 
				\multirow{8}{*}{\rotatebox[origin=c]{90}{\hspace{-10pt}\parbox{1.1cm}{\centering\tiny\textbf{Visual Genome}}}}
				& \textsc{Freq}~\cite{zellers2018neural} & $-$ & 45.4 & 0.5 & 51.7 & 18.3 & 19.1 & & 69.8 & 0.3 & 89.8 & 31.0 & 22.1 \Tstrut \Bstrut\\ 
				\cline{2-14}
				& \multirow{2}{*}{MP~\cite{xu2017scene,zellers2018neural}}
				& \textsc{Baseline~\eqref{eq:loss_baseline}} & 47.2 & 8.2 & 51.9 & 26.2 & \cellcolor{extreme}17.3 & & 74.8 & \cellcolor{bad}23.3 & 86.6 & \cellcolor{bad}51.3 & \cellcolor{extreme}20.6 \Tstrut \\
				& & \textsc{Ours~\eqref{eq:loss_norm}} & 48.6 & \textbf{9.1} & \textbf{52.6} & \textbf{28.2} & \cellcolor{extreme}\textbf{26.5} & & 78.2 & \cellcolor{bad}\textbf{28.4} & 89.4 & \cellcolor{bad}58.4 & \cellcolor{extreme}32.1 \Bstrut \\
				\cline{2-14}
				& \multirow{3}{*}{NM~\cite{zellers2018neural}} & \textsc{Baseline~\eqref{eq:loss_baseline}} & 48.1 & \cellcolor{bad}5.7 & 51.9 & 26.5 & \cellcolor{bad}20.4 & & 80.5 & \cellcolor{extreme}11.1 & 91.0 & \cellcolor{bad}51.8 & \cellcolor{bad}26.9\Tstrut\\
				& & \textsc{Ours~\eqref{eq:loss_norm}} & {48.4} & \cellcolor{bad}{7.1} & {52.0} & {27.7} & \cellcolor{bad}25.5 & & 82.0 & \cellcolor{extreme}{16.7} & {92.0} & \cellcolor{bad}{56.4} & \cellcolor{bad}34.8 \\
				& & \textsc{Ours~\eqref{eq:loss_norm}, no Freq} & {48.4} & \cellcolor{bad}{8.9} & 51.8 & {28.0} & \cellcolor{bad}26.1 & & {82.5} & \cellcolor{extreme}{26.6} & \textbf{{92.4}} & \cellcolor{bad}{\textbf{60.3}} & \cellcolor{bad}35.8\Bstrut \\
				\cline{2-14}
				& KERN$^\star$~\cite{chen2019knowledge} & \textsc{Baseline~\eqref{eq:loss_baseline}} & 49.0 & 3.7 & \textbf{52.6} & 27.7 & 26.2 & & 81.9 & 5.8 & 91.9 & 49.1 & \textbf{36.3}\Tstrut\\
				& RelDN$^\star$~\cite{zhang2019graphical} & \textsc{Baseline~\eqref{eq:loss_baseline}} & \textbf{50.8$^\dagger$} & $-$ & $-$ & $-$ & $-$ & & \textbf{93.7$^\dagger$} & $-$ & $-$ & $-$ & $-$\Bstrut\\
				\Xhline{2\arrayrulewidth}
				\multirow{2}{*}{\rotatebox[origin=c]{90}{\parbox{0.48cm}{\tiny\vspace{2pt}\textbf{GQA}}}} & \multirow{2}{*}{MP~\cite{xu2017scene,zellers2018neural}} & \textsc{Baseline~\eqref{eq:loss_baseline}} & 27.1 & 2.8 & 31.9 & \textbf{{8.9}} & \cellcolor{extreme}1.6 & & 59.7 & 34.9 & 96.4 & 88.4 & \cellcolor{extreme}1.8 \Tstrut\\
				\multirow{3}{*}{\rotatebox[origin=c]{90}{\parbox{1.4cm}{\tiny\vspace{0pt}\centering\hspace{-1pt}\textbf{GQA\\-nLR}}}} & & \textsc{Ours~\eqref{eq:loss_norm}} & \textbf{27.6} & \textbf{{3.0}} & \textbf{{32.2}} & \textbf{{8.9}} & \cellcolor{extreme}\textbf{2.8} & & \textbf{{61.0}} & {\textbf{37.2}} & {\textbf{96.9}} & \textbf{89.5} & \cellcolor{extreme}\textbf{2.9} \Bstrut \\
				\hline
				& \multirow{2}{*}{MP~\cite{xu2017scene,zellers2018neural}} & \textsc{Baseline~\eqref{eq:loss_baseline}} & 24.9 & 3.0 & \textbf{30.2} & 12.4 & \cellcolor{extreme}2.8 & & 58.1 & \cellcolor{bad}21.7 & 71.6 & \cellcolor{bad}47.0 & \cellcolor{extreme}4.6 \Tstrut\\
				& & \textsc{Ours~\eqref{eq:loss_norm}} & \textbf{25.0} & \textbf{3.2} & 29.4 & \textbf{12.6} & \cellcolor{extreme}\textbf{7.0} & & \textbf{62.4} & \cellcolor{bad}\textbf{26.2} & \textbf{77.9} & \cellcolor{bad}\textbf{55.0} & \cellcolor{extreme}\textbf{12.1} \\
			\end{tabular}
		\end{center}
		\vspace{-10pt}
		\caption{\small Results on Visual Genome (split~\cite{xu2017scene}) and GQA~\cite{hudson2019gqa}. We obtain particularly strong results in columns R$_{ZS}$, wR$_{tr}$ and mR in each of the two tasks. \crule[bad]{12pt}{8pt} denotes cases with $\geq 15\%$ relative difference between the baseline and our result; \crule[extreme]{12pt}{8pt} denotes a difference of $\geq 50\%$. Best results for each dataset (VG, GQA and GQA-nLR) are bolded.
			GQA-nLR: our version of GQA with left/right spatial relationships excluded, where scene graphs become much sparser (see~\apdx~for dataset details). $^\star$Results are provided for the reference and evaluating our loss with these methods is left for future work. 
			$^\dagger$The correctness of this evaluation is discussed in~\cite{tang2020github}.}\label{table:main_results}
	\end{table}
	
	\vspace{-3pt}
	\paragraph{VG results.} We can observe that both Message Passing (MP) and Neural Motifs (NM) greatly benefit from our density-normalized loss on all reported metrics. Larger gaps are achieved on metrics evaluating zero and few-shots. For example, in PredCls on Visual Genome, MP with our loss is 22\% better (in relative terms) on zero-shots, while NM with our loss is 50\% better. The gains arising from other zero-shot and weighted metrics are also significant.
	
	\vspace{-3pt}
	\paragraph{GQA results.} On GQA, our loss also consistently improves results, especially in PredCls. However, the gap is lower compared to VG. There are two reasons for this: 1) scene graphs in GQA are much denser (see~\apdx), i.e. the imbalance between FG and BG edges is less pronounced, which means that in the baseline loss the edge term is not diminished to the extent it is in VG; and 2) the training set of GQA is more diverse than VG (with 15 times more labeled triplets), which makes the baseline model generalize well on zero and few-shots. We confirm these arguments by training and evaluating on our version of GQA: \textbf{GQA-nLR} with left and right predicate classes excluded making scene graph properties, in particular sparsity, more similar to those of VG.
	\looseness-1

	\begin{wrapfigure}{r}{7cm}
		\vspace{-25pt}
		\centering
		\includegraphics[width=6cm,align=c,trim={0.2cm 0 0 0},clip]{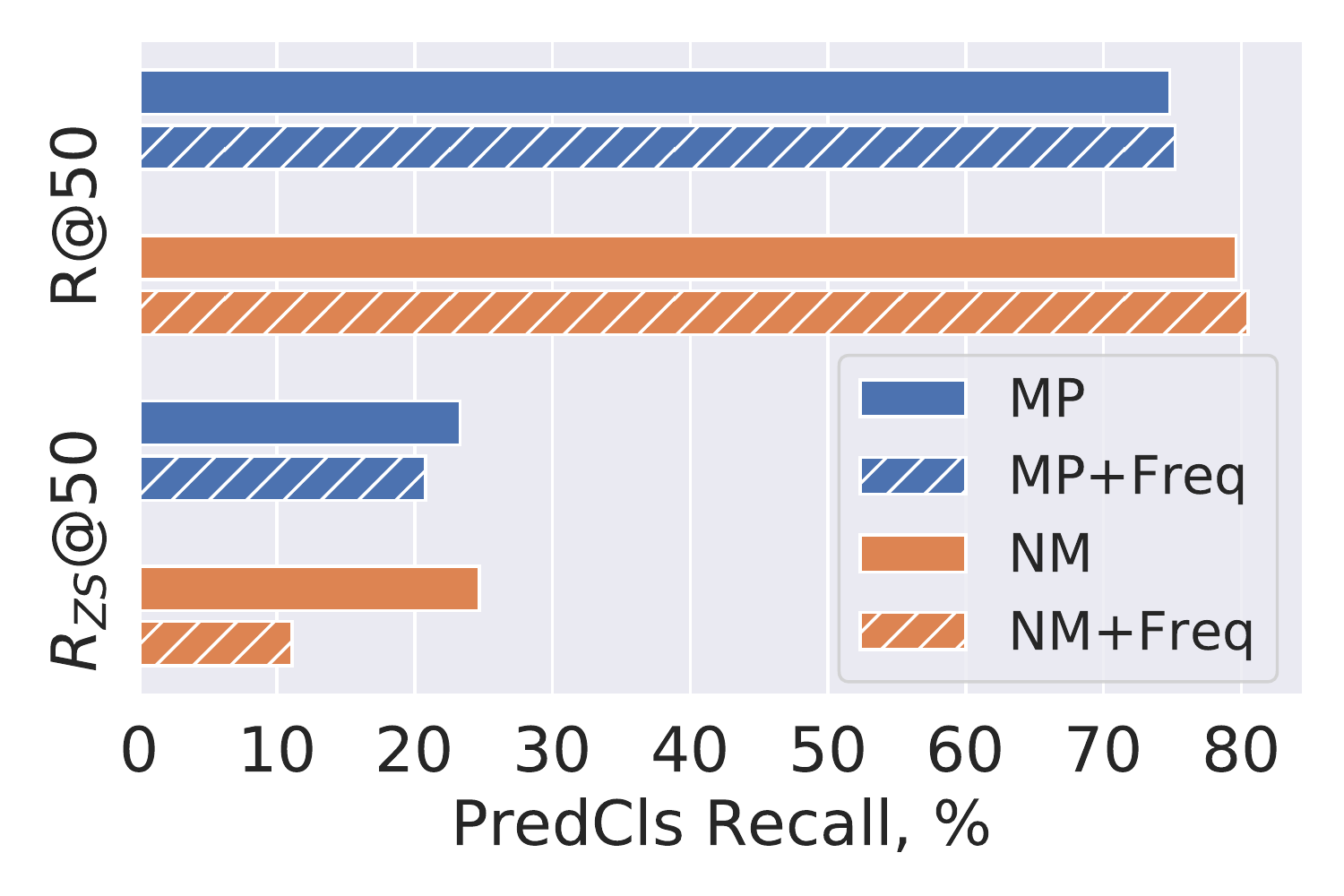}
		\vspace{-10pt}
		\caption{\small Ablating \textsc{Freq}. \textsc{Freq} only marginally improves results on R@50. At the same time, it leads to large drops in zero-shot recall R$_{ZS}$@50 (and our weighted triplet recall, see Table~\ref{table:main_results}).}\label{fig:freq}
		\vspace{-5pt}
	\end{wrapfigure}
	
	\vspace{-3pt}
	\paragraph{Effect of the Frequency Bias (\textsc{Freq}) on Zero and Few-Shot Performance.}
	The \textsc{Freq} model~\cite{zellers2018neural} simply predicts the most frequent predicate between a subject and an object, $P(R|O)$.
	Its effect on few-shot generalization has not been empirically studied before. 
	We study this by adding/ablating \textsc{Freq} from baseline MP and NM on Visual Genome (Figure~\ref{fig:freq}). Our results indicate that \textsc{Freq} only marginally improves results on unweighted metrics. At the same time, perhaps unsurprisingly, it leads to severe drops in zero-shot and weighted metrics, especially in NM. For example, by ablating \textsc{Freq} from NM, we improve PredCls-R$_{ZS}$@50 from 11\% to 25\%. This also highlights that the existing recall metrics are a poor choice to understand the effectiveness of a model.\looseness-1
	
	\begin{wrapfigure}{r}{7cm}
		\vspace{-5pt}
		\centering
		\includegraphics[width=6cm,align=c,trim={0.2cm 0 0 0},clip]{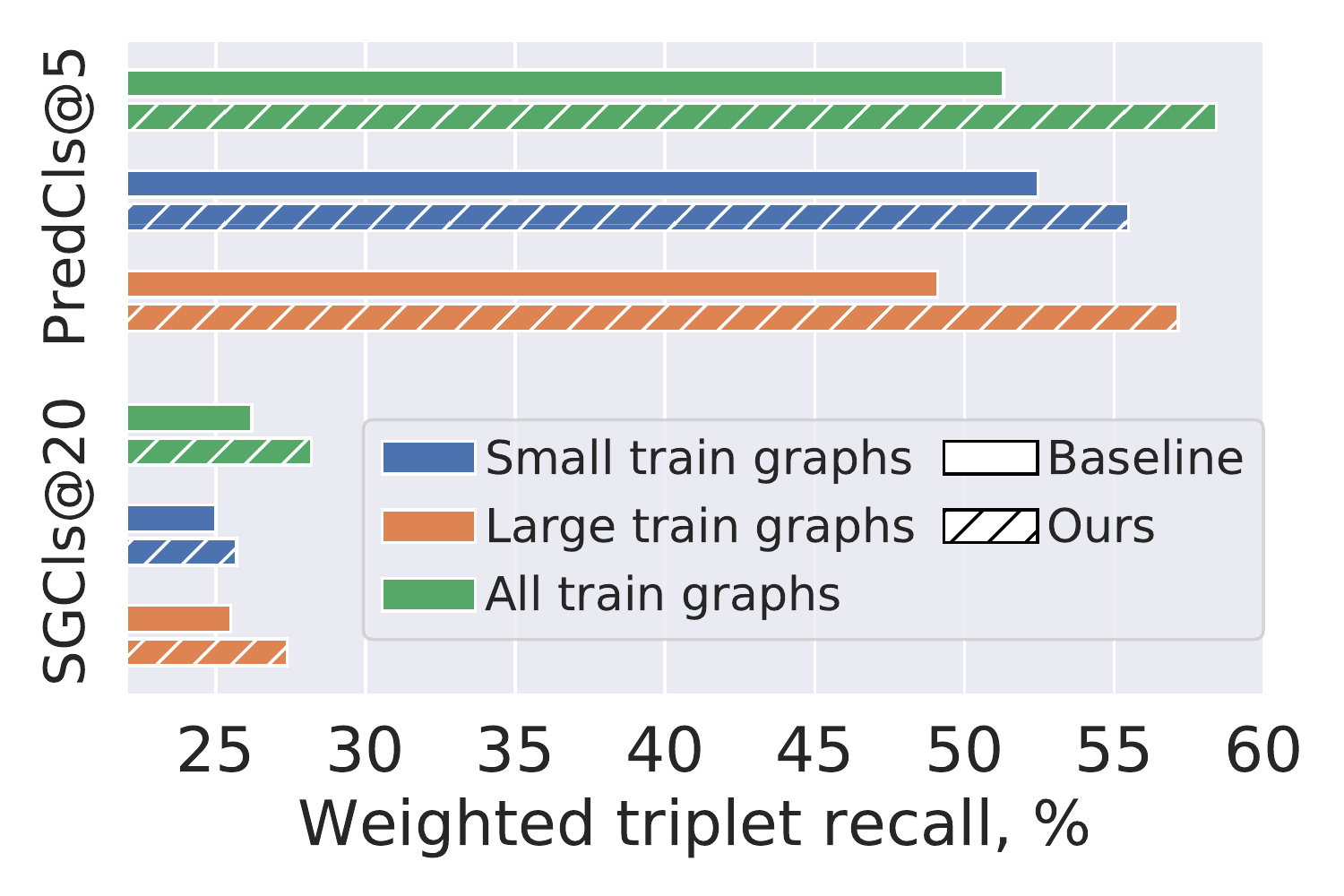}
		\vspace{-10pt}
		\caption{\small Learning from small ($N \leq 10$) \textit{vs.}~large ($N>10$) graphs. Our loss makes models learn from larger graphs more effectively, which is important for generalization, because such graphs contain a lot of labels (see Figure~\ref{fig:motivation}).}\label{fig:results_graph_size}
		\vspace{-5pt}
	\end{wrapfigure}
	
	\paragraph{Why does loss normalization help more on few and zero-shots?}
	The baseline loss effectively ignores edge labels of large graphs, because it is scaled by a small $d$ in those cases (Figure~\ref{fig:graph_density}). To validate that, we split the training set of Visual Genome in two subsets, with a comparable number of images in each: with relatively small and large graphs evaluating on the original test set in both cases. We observe that the baseline model does not learn well from large graphs, while our loss enables this learning (Figure~\ref{fig:results_graph_size}). Moreover, when trained on small graphs only, the baseline is even better in PredCls than when trained on all graphs. This is because in the latter case, large graphs, when present in a batch, make the whole batch more sparse, downweighting the edge loss of small graphs as well. 
	At the same time, larger graphs predictably contain more labels, including many few-shot labels (Figure~\ref{fig:motivation}). %
	Together, these two factors make the baseline ignore many few-shot triplets pertaining to larger graphs at training time, so the model cannot generalize to them at test time.
	Since the baseline essentially observes less variability during training, it leads to poor generalization on zero-shots as well. This argument aligns well to the works from other domains~\cite{hill2019environmental}, showing that generalization strongly depends on the diversity of samples during training.
	Our loss fixes the issue of learning from larger graphs, which, given the reasons above, directly affects the ability to generalize.\looseness-1

	\begin{table}[]
		\scriptsize
		\begin{center}
			\setlength{\tabcolsep}{2pt}
			\begin{minipage}{.46\linewidth}
				\centering
				\begin{tabular}{lcc}
					\textbf{Model} & \textbf{SGCls-GC} & \textbf{PredCls-GC} \\ 
					\noalign{\smallskip}
					\Xhline{2\arrayrulewidth}
					\noalign{\smallskip}
					VTE~\cite{zhang2017visual, wang2019generating} & $-$ & 16.4 \\
					STA~\cite{yang2018shuffle} & $-$ & 18.9 \\
					ST-GAN~\cite{wang2019generating} & $-$ & 19.0 \\
					MP, baseline~\eqref{eq:loss_baseline} & 2.3 & 20.4 \\
					\noalign{\smallskip}
					\hline
					\noalign{\smallskip}
					MP, ours~\eqref{eq:loss_norm} & \textbf{3.1} & \textbf{21.4} \\
				\end{tabular}
				\caption{\small{Zero-shot results (R$_{ZS}$@100) on the VTE split~\cite{zhang2017visual}.}}\label{table:zs_comparison_vte}
			\end{minipage}%
			\hspace{10pt}
			\begin{minipage}{.46\linewidth}
				\centering
				\begin{tabular}{lcc}
					\textbf{Model} & \textbf{SGCls-GC} & \textbf{PredCls-GC} \\ 
					\noalign{\smallskip}
					\Xhline{2\arrayrulewidth}
					\noalign{\smallskip}
					NM+TDE~\cite{tang2020unbiased} & \textbf{4.5} & 18.2 \\
					NM, baseline~\eqref{eq:loss_baseline} & 1.7 & 9.5 \\
					MP, baseline~\eqref{eq:loss_baseline} & 3.2 & 20.1 \\
					\noalign{\smallskip}
					\hline
					\noalign{\smallskip}
					NM, ours~\eqref{eq:loss_norm}, no Freq & 3.9 & 20.4 \\
					MP, ours~\eqref{eq:loss_norm} & 4.2 & \textbf{21.5} \\
				\end{tabular}
				\caption{\small {Zero-shot results (R$_{ZS}$@100) on the VG split~\cite{xu2017scene}. \citet{tang2020unbiased} uses ResNeXt-101 as a backbone, which helps to improve results.}}\label{table:zs_comparison_vg}
			\end{minipage}
			\smallskip \smallskip \smallskip \smallskip \\
			\setlength{\tabcolsep}{3pt}
			\begin{tabular}{l|ll|cc|cc}
				\multicolumn{3}{c|}{} & \multicolumn{2}{c|}{\textbf{Testing on VG}} & \multicolumn{2}{c}{\textbf{Testing on GQA}} \Bstrut \\
				\cline{4-7}
				\multirow{1}{*}{\textbf{Tuning dataset}} &
				\multirow{1}{*}{\textbf{Hyperparams}} & 
				\multirow{1}{*}{\textbf{Loss}} & \multicolumn{1}{c}{{SGCls@100}} & \multicolumn{1}{c|}{{PredCls@50}} & \multicolumn{1}{c}{{SGCls@100}} & \multicolumn{1}{c}{{PredCls@50}} \Tstrut  \Bstrut\\
				\Xhline{2\arrayrulewidth}
				No tune (baseline) & $-$ & \eqref{eq:loss_baseline} & 47.2/8.2 & 74.8/23.3 & 27.1/2.8 & 59.7/34.9 \Tstrut \Bstrut\\
				\hline
				VG & $\lambda=20$ & \eqref{eq:loss_tune2} & 48.9/9.2 & \textbf{78.3}/27.9 & 26.6/2.6 & 60.4/36.9 \Tstrut \\
				VG & $\alpha=0.5, \beta=20$ & \eqref{eq:loss_tune1} & \textbf{49.1}/9.4 & 78.2/27.8 & 27.1/2.9 & 60.5/36.3  \\
				GQA & $\lambda=5$ & \eqref{eq:loss_tune2} & 48.8/9.2 & 78.0/26.8 & \textbf{27.8}/2.9 & 60.5/36.1 \\
				GQA & $\alpha=1, \beta=5$ & \eqref{eq:loss_tune1} & 48.6/8.7 & 77.4/27.8 & 27.5/2.9 & 60.7/36.6 \Bstrut \\
				\hline
				No tune (ours, independ. norm) & $\alpha=\beta=1$ & \eqref{eq:loss_tune1} & 47.5/8.4 & 74.3/25.3 & 27.4/2.9 & 59.5/35.4 \Tstrut\\
				No tune (ours, no upweight) & $\gamma=0.05/0.2$ for VG/GQA & \eqref{eq:loss_norm} & 48.7/\textbf{9.6} & \textbf{78.3}/28.2 & 27.4/2.9 & \textbf{61.1}/36.8 \\
				No tune (ours) & $\gamma=1$ & \eqref{eq:loss_norm} & 48.6/9.1 & 78.2/\textbf{28.4} & 27.6/\textbf{3.0} & 61.0/\textbf{37.2} \\
			\end{tabular}
		\end{center}
		\vspace{-5pt}
		\caption{\small Comparing our loss to other approaches using MP~\cite{xu2017scene,zellers2018neural} and R/R$_{ZS}$@K metrics. Best results for each metric are bolded.}\label{table:loss_ablations}
	\end{table}
	
	\paragraph{Alternative approaches.}
	We compare our loss to ones with tuned hyperparameters $\alpha,\beta,\lambda$ (Table~\ref{table:loss_ablations}):
	\begin{align}
	{\cal L} = {\cal L}_{node} + &\alpha {\cal L}_{FG}  + \beta {\cal L}_{BG},\label{eq:loss_tune1} \\
	{\cal L} = {\cal L}_{node} + &\lambda {\cal L}_{edge}.\label{eq:loss_tune2}
	\end{align}
	Our main finding is that, while these losses can give similar or better results in some cases, the parameters $\alpha$, $\beta$ and $\lambda$ do not generally transfer across datasets and must be tuned every time, which can be problematic at larger scale~\cite{zhang2019large}. In contrast, our loss does not require tuning and achieves comparable performance.

	To study the effect of density normalization separately from upweighting the edge loss (which is a side effect of our normalization), we also consider downweighting our edge term~\eqref{eq:loss_norm} by some $\gamma<1$ to cancel out this upweighting effect.
	This ensures a similar range for the losses in our comparison.
	We found (Table~\ref{table:loss_ablations}) that the results are still significantly better than the baseline and, in some cases, even better than our hyperparameter-free loss. This further confirms that normalization of the graph density is important on its own. When carefully fine-tuned, the effects of normalization and upweighting are complimentary (\eg~when $\alpha,\beta$ or $\gamma$ are fine-tuned, the results tend to be better).
	
	\vspace{-6pt}
	\paragraph{Comparison to other zero-shot works.}
	We also compare to previous works studying zero-shot generalization (Tables~\ref{table:zs_comparison_vte} and~\ref{table:zs_comparison_vg}). %
	For comprehensive evaluation, we test on both VTE and VG splits. We achieve superior results on VTE, even by just using the baseline MP, because, as shown in our main results, it generalizes well. On the VG split, we obtain results that compete with a more recent Total Direct Effect (TDE) method~\cite{tang2020unbiased}, even though the latter uses a more advanced detector and feature extractor.
	In all cases, our loss improves baseline results and, except for R$_{ZS}$@100 in SGCls, leads to state-of-the-art generalization.
	Our loss and TDE can be applied to a wide range of models, beyond MP and NM, to potentially have a complementary effect on generalization, which is interesting to study in future work.
	
	\setlength{\tabcolsep}{4pt}
	\begin{table}[]
		\vspace{-7pt}
		\scriptsize
		\begin{center}
			\begin{tabular}{llllccccccc} %
				& \multirow{2}{*}{\textbf{Model}} & \multirow{2}{*}{\textbf{Backbone}} & \multirow{2}{*}{\textbf{Loss}} & \multicolumn{3}{c}{\textbf{SGGen-GC}} & &  \multicolumn{3}{c}{\textbf{SGGen}} \Bstrut \\
				\cline{5-7}\cline{9-11}
				& & & & \tiny R@100 & \tiny R$_{ZS}$@100 & \tiny mR@100 & 
				& \tiny R@100 & \tiny R$_{ZS}$@100 & \tiny mR@100 \Tstrut \Bstrut\\
				\Xhline{2\arrayrulewidth}
				& 
				\textsc{Freq}~\cite{zellers2018neural} & VGG16 & & 27.6 & 0.02 & 5.6 & & 30.9 & 0.1 & 8.9\Tstrut\Bstrut \\
				\cline{2-11}
				& \multirow{3}{*}{MP~\cite{xu2017scene,zellers2018neural}} & VGG16 & \textsc{Baseline~\eqref{eq:loss_baseline}} & 24.3 & 0.8 & \cellcolor{bad}4.5 & & 27.2 & \cellcolor{bad}0.9 & \cellcolor{bad}7.1\Tstrut \\
				& & VGG16 & \textsc{Ours~\eqref{eq:loss_norm}} & 25.2 & 0.9 & \cellcolor{bad}5.8 & & 28.2 & \cellcolor{bad}1.2 & \cellcolor{bad}9.5\Bstrut\\
				
				& \multirow{3}{*}{NM~\cite{zellers2018neural}} & VGG16 & \textsc{Baseline~\eqref{eq:loss_baseline}} & 29.8 & \cellcolor{extreme}0.3 & \cellcolor{bad}5.9 & & 35.0 & \cellcolor{extreme}0.8 & \cellcolor{bad}12.4\Tstrut \\
				& & VGG16 & \textsc{Ours~\eqref{eq:loss_norm}} & 29.4 & \cellcolor{extreme}1.0 & \cellcolor{bad}\textbf{8.1} & & 35.0 & \cellcolor{extreme}1.8 & \cellcolor{bad}15.4 \\
				& & VGG16 & \textsc{Ours~\eqref{eq:loss_norm}, no Freq} & {30.4} & \cellcolor{extreme}\textbf{1.7} & \cellcolor{bad}7.8 & & {35.9} & \cellcolor{extreme}\textbf{2.4} & \cellcolor{bad}15.3\Bstrut \\
				& KERN~\cite{chen2019knowledge} & VGG16 & \textsc{Baseline~\eqref{eq:loss_baseline}} & 29.8 & 0.04 & 7.3 & & 35.8 & 0.02 & \textbf{16.0}\Tstrut\\
				& RelDN~\cite{zhang2019graphical} & VGG16 & \textsc{Baseline~\eqref{eq:loss_baseline}} & \textbf{32.7} & $-$ & $-$ & & \textbf{ 36.7} & $-$ & $-$\Bstrut \\
				\cline{2-11}
				& RelDN~\cite{zhang2019graphical} & ResNeXt-101 & \textsc{Baseline~\eqref{eq:loss_baseline}} & 36.7 & $-$ & $-$ & & 40.0 & $-$ & $-$\Tstrut \\
				& NM~\cite{tang2020unbiased} & ResNeXt-101 & \textsc{Baseline~\eqref{eq:loss_baseline}} & \textbf{36.9} & 0.2 & 6.8 & & $-$ & $-$ & $-$ \\
				& NM+TDE~\cite{tang2020unbiased} & ResNeXt-101 & \textsc{Baseline~\eqref{eq:loss_baseline}} & 20.3 & 2.9 & 9.8 & & $-$ & $-$ & $-$  \\
				& VCTree+TDE~\cite{tang2020unbiased} & ResNeXt-101 & \textsc{Baseline~\eqref{eq:loss_baseline}} & 23.2 & \textbf{3.2} & \textbf{11.1} & & $-$ & $-$ & $-$\Bstrut \\
			\end{tabular}
		\end{center}
		\vspace{-15pt}
		\caption{\small Comparison of our SGG methods to the state-of-the-art on Visual Genome (split~\cite{xu2017scene}). We report results with and without the graph constraint, \textbf{SGGen-GC} and \textbf{SGGen} respectively. %
			All models use Faster R-CNN~\cite{ren2015faster} as a detector, but the models we evaluate use a weaker backbone compared to~\cite{tang2020unbiased}. Interestingly, TDE's improvement on zero-shots (R$_{ZS}$) and mean recall (mR) comes at a significant drop in R@100, which means that frequent triplets are recognized less accurately. Our loss does not suffer from this. Table cells are colored the same way as in Table~\ref{table:main_results}. %
		}\label{table:sggen_results}
		\vspace{-6pt}
	\end{table}
	
	\vspace{-6pt}
	\paragraph{Comparison on SGGen, $P ({\cal G}|I)$.}
	In SGCls and PredCls, we relied on ground truth bounding boxes $B_{gt}$, while in SGGen the bounding boxes $B_{pred}$ predicted by a detector should be used to enable a complete image-to-scene graph pipeline. Here, even small differences between $B_{gt}$ and $B_{pred}$ can create large distribution shifts between corresponding extracted features $(V,E)$ (see Section~\ref{sec:baseline}), on which SGCls models are trained. Therefore, it is important to \textit{refine} the SGCls model on $(V,E)$ extracted based on predicted $B_{pred}$, according to previous work~\cite{zellers2018neural,chen2019knowledge}. In our experience, this refinement can boost the R@100 result by around 3\% for Message Passing and up to 8\% for Neural Motifs (in absolute terms). In Table~\ref{table:sggen_results}, we report results after the refinement completed both for the baseline loss and our loss in the same way. Similarly to the SGCls and PredCls results, our loss consistently improves baseline results in SGGen. It also allows Neural Motifs (NM) to significantly outperform KERN~\cite{chen2019knowledge} on zero-shots (R$_{ZS}$@100), while being only slightly worse in one of the mR@100 results. The main drawback of KERN is its slow training, which prevented us to explore this model together with our loss. 
	Following our experiments in Table~\ref{table:main_results} and Figure~\ref{fig:freq}, we also confirm the positive effect of removing \textsc{Freq} from NM.
	A more recent work of~\citet{tang2020unbiased} shows better results on zero-shots and mean recall, however, we note their more advanced feature extractor, therefore it is difficult to compare our results to theirs in a fair fashion. But, since they also use the baseline loss~\eqref{eq:loss_baseline}, our loss~\eqref{eq:loss_norm} can potentially improve their model, which we leave for future work. 
	
	\begin{wraptable}{r}{6.3cm}
		\small
		\vspace{-10pt}
		\setlength{\tabcolsep}{2pt}
		\begin{tabular}{lccc} 
			\Xhline{2\arrayrulewidth} 
			\textbf{Loss} & \footnotesize R@300 & \footnotesize R$_{ZS}$@300 &  \footnotesize mR$_{tr}$@300\Bstrut\Tstrut\\
			\Xhline{2\arrayrulewidth} 
			\textsc{Baseline}~\eqref{eq:loss_baseline} & 6.2 & 0.5 & 1.3 \Tstrut \\ 
			\textsc{Ours}~\eqref{eq:loss_norm}  & \textbf{6.3} & \textbf{0.7} & \textbf{2.4} \Bstrut\\ 
		\end{tabular}
		\vspace{-5pt}
		\caption{\small \textbf{SGGen} results on GQA~\cite{hudson2019gqa} using MP. Mask R-CNN~\cite{he2017mask} fine-tuned on GQA is used in this task.}\label{table:gqa_sgg}
	\end{wraptable} 
	
	Finally, we evaluate SGGen on GQA using Message Passing (Table~\ref{table:gqa_sgg}), where we also obtain improvements with our loss. GQA has 1703 object classes compared to 150 in VG making object detection harder. When evaluating SGGen, the predicted triplet is matched to ground truth (GT) if predicted and GT bounding boxes have an intersection over union (IoU) of $\geq$50\%, so more misdetections lead to a larger gap between SGCls and SGGen results.\looseness-1

	\begin{figure}
		\vspace{-10pt}
		\newcommand\Tstrutmore{\rule{0pt}{10ex}}         %
		\newcommand\Bstrutmore{\rule[-2.3ex]{0pt}{0pt}}
		\newcommand{\figwidthvis}{0.16\textwidth}
		\centering
		\scriptsize
		\setlength{\tabcolsep}{0pt}
		\begin{tabular}{p{0.3cm}|rc|p{0.1cm}rc|p{0.1cm}rc}
			& \multicolumn{2}{c|}{\textsc{Ground truth}} & & \multicolumn{2}{c|}{\textsc{Baseline}} & & \multicolumn{2}{c}{\textsc{Ours}} \\
			& \textsc{Detections} & \textsc{Scene Graph} & & \textsc{Detections} & \textsc{Scene Graph} & & \textsc{Detections} & \textsc{Scene Graph} \\
			\Xhline{5\arrayrulewidth}
			\multirow{8}{*}{\rotatebox[origin=c]{90}{\footnotesize\parbox{6cm}{\vspace{0pt}\textbf{Baseline is correct, Ours is \textcolor{red}{\textbf{incorrect}}}}}} &  \includegraphics[width=\figwidthvis, align=c]{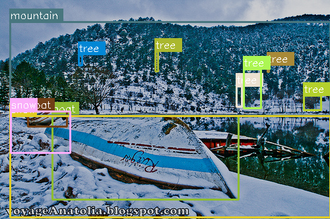} &
			\includegraphics[width=\figwidthvis, align=c]{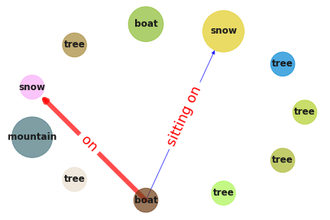} & & \includegraphics[width=\figwidthvis, align=c]{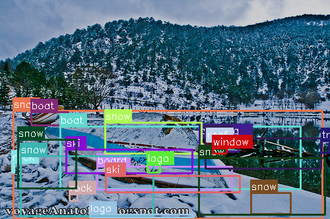} &
			\includegraphics[width=\figwidthvis, align=c]{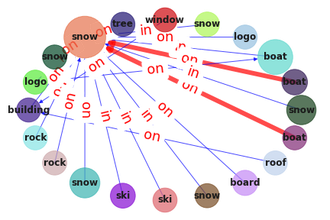} & 
			& \includegraphics[width=\figwidthvis, align=c]{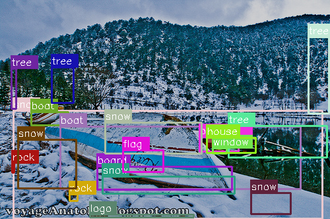} & 
			\includegraphics[width=\figwidthvis, align=c]{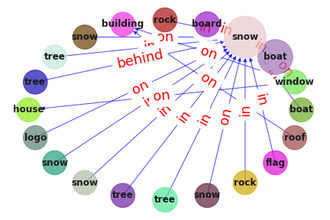} \Tstrutmore\Bstrut\\
			& zero-shot triplet: & {boat on snow} & & match: & boat on snow & & closest match: & {boat \textcolor{red}{\textbf{in}} snow} \Bstrut\\
			\cline{2-9} 
			& \includegraphics[width=\figwidthvis, align=c]{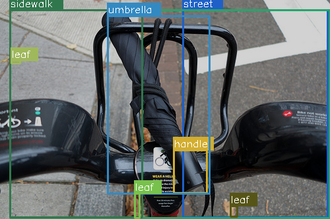} &
			\includegraphics[width=\figwidthvis, align=c]{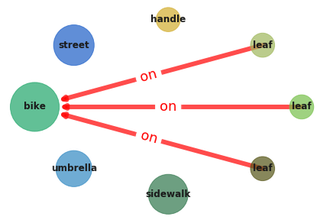} & & \includegraphics[width=\figwidthvis, align=c]{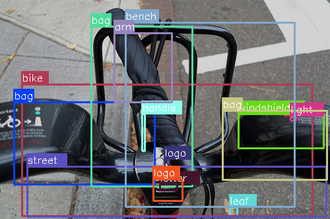} &
			\includegraphics[width=\figwidthvis, align=c]{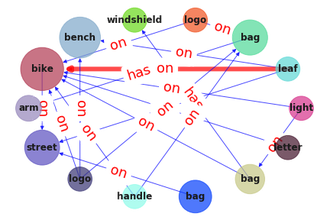} & 
			& \includegraphics[width=\figwidthvis, align=c]{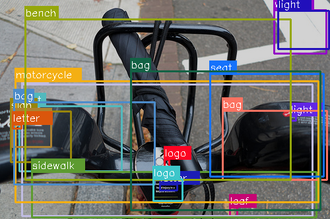} & 
			\includegraphics[width=\figwidthvis, align=c]{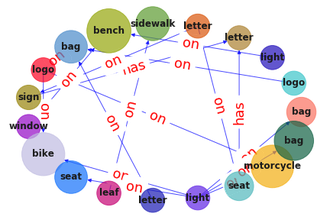} \Tstrutmore\Bstrut\\
			& \multicolumn{2}{r|}{zero-shot triplet: \textcolor{violet}{\textbf{leaf on bike}} (mislabeled)} & & match: & leaf on bike & & closest match: & {leaf on \textcolor{red}{\textbf{sidewalk}}} \Bstrut\\
			\cline{2-9} 
			& \includegraphics[width=\figwidthvis, align=c]{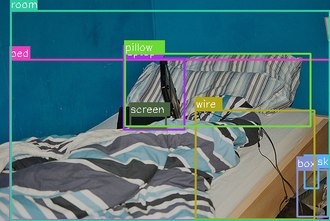} &
			\includegraphics[width=\figwidthvis, align=c]{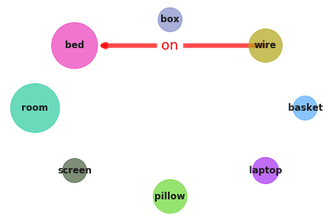} & & \includegraphics[width=\figwidthvis, align=c]{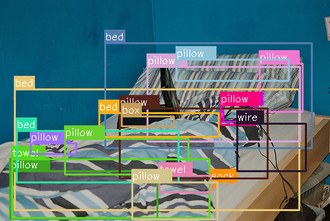} &
			\includegraphics[width=\figwidthvis, align=c]{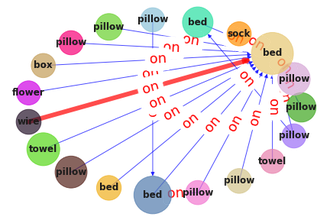} & 
			& \includegraphics[width=\figwidthvis, align=c]{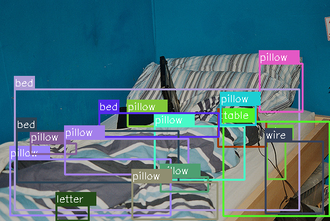} & 
			\includegraphics[width=\figwidthvis, align=c]{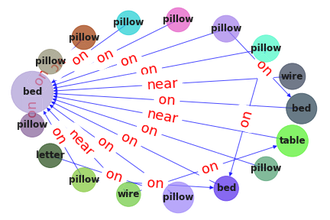} \Tstrutmore\Bstrut\\
			& zero-shot triplet: & {wire on bed} & & match: & wire on bed & & closest match: & {wire \textcolor{red}{\textbf{near}} bed} \Bstrut\\
			\cline{2-9} 
			& \includegraphics[width=\figwidthvis, height=1.8cm, align=c]{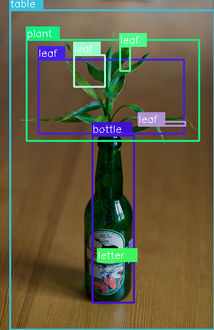} &
			\includegraphics[width=\figwidthvis, align=c]{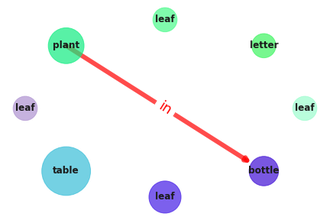} & 
			& \includegraphics[width=\figwidthvis, height=1.8cm, align=c]{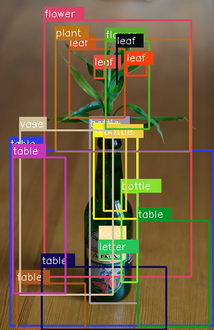} &
			\includegraphics[width=\figwidthvis, align=c]{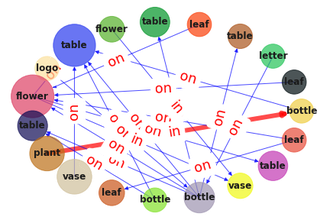} & 
			& \includegraphics[width=\figwidthvis, height=1.8cm, align=c]{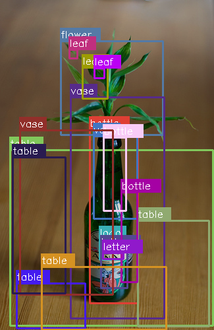} &
			\includegraphics[width=\figwidthvis, align=c]{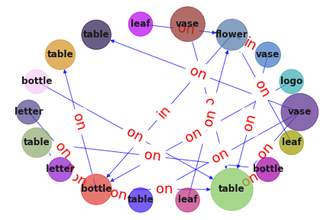} \Tstrutmore\Bstrut\\
			& zero-shot triplet: & {plant in bottle} & & match: & plant in bottle & & closest match: & {\textcolor{red}{\textbf{flower}} in bottle}\Bstrut\\
			
			\Xhline{5\arrayrulewidth}
			\multirow{8}{*}{\rotatebox[origin=c]{90}{\footnotesize\parbox{6cm}{\textbf{Baseline is \textcolor{red}{\textbf{incorrect}}, Ours is correct}}}} &
			\includegraphics[width=\figwidthvis, height=1.8cm, align=c]{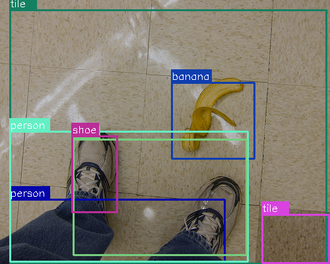} &
			\includegraphics[width=\figwidthvis, align=c]{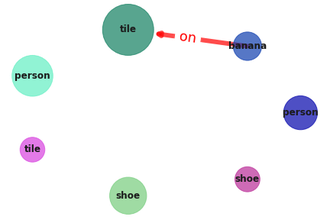} & &  \includegraphics[width=\figwidthvis, height=1.8cm, align=c]{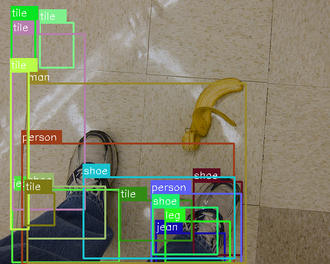} &
			\includegraphics[width=\figwidthvis, align=c]{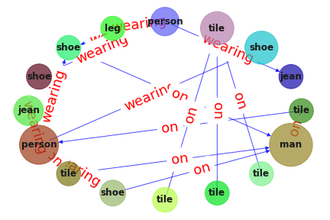} & 
			& 
			\includegraphics[width=\figwidthvis, height=1.8cm, align=c]{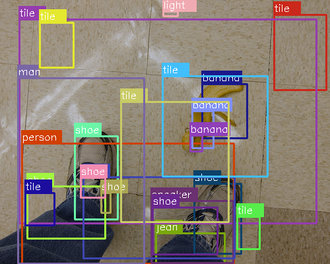} & 
			\includegraphics[width=\figwidthvis, align=c]{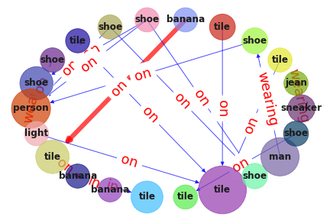} \Tstrutmore\Bstrut\\
			& zero-shot triplet: & {banana on tile} & & \multicolumn{2}{r|}{no triplet involving a banana in top-20} & & match: & {banana on tile} \Bstrut\\
			\cline{2-9} 
			& \includegraphics[width=\figwidthvis, height=1.8cm, align=c]{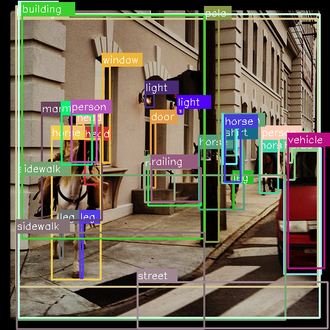} &
			\includegraphics[width=\figwidthvis, align=c]{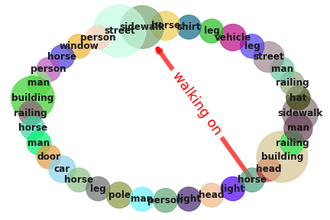} & & \includegraphics[width=\figwidthvis, height=1.8cm, align=c]{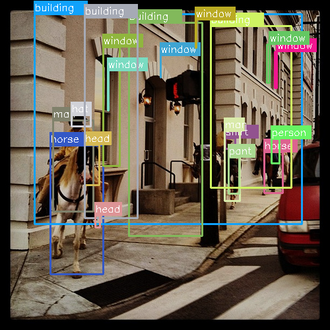} &
			\includegraphics[width=\figwidthvis, align=c]{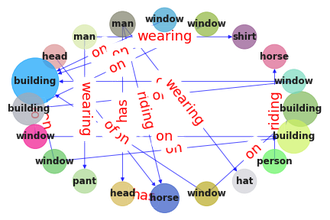} & 
			& \includegraphics[width=\figwidthvis, height=1.8cm, align=c]{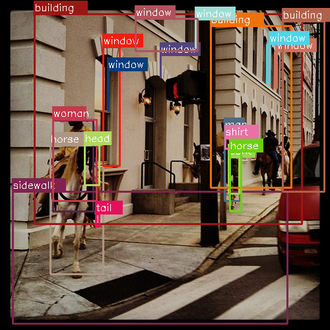} & 
			\includegraphics[width=\figwidthvis,align=c]{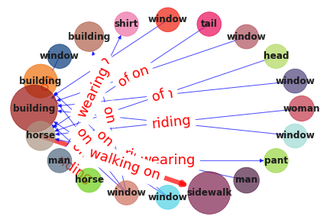} \Tstrutmore\Bstrut\\
			& \multicolumn{2}{r|}{zero-shot triplet: horse walking on sidewalk} & & \multicolumn{2}{r|}{no triplet involving a sidewalk in top-20} & & \multicolumn{2}{r}{match: horse walking on sidewalk} \Bstrut\\
			\cline{2-9}  
			& \includegraphics[width=\figwidthvis, align=c]{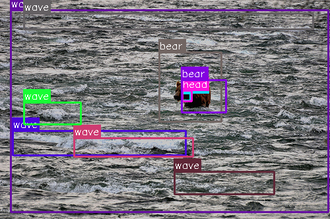} &
			\includegraphics[width=\figwidthvis, align=c]{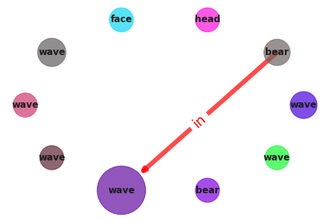} & & \includegraphics[width=\figwidthvis, align=c]{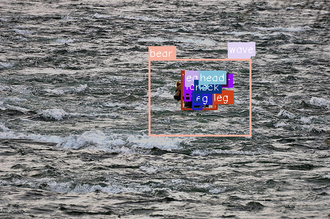} &
			\includegraphics[width=\figwidthvis, align=c]{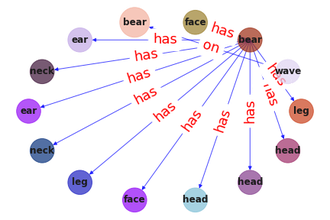} & 
			& \includegraphics[width=\figwidthvis, align=c]{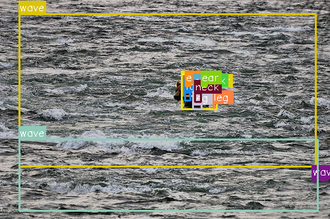} & 
			\includegraphics[width=\figwidthvis, align=c]{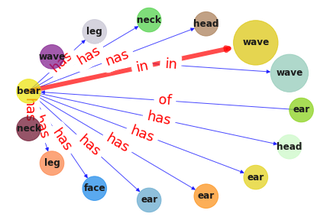} \Tstrutmore\Bstrut\\
			& zero-shot triplet: & {bear in wave} & & closest match: & bear \textcolor{red}{\textbf{on}} wave & & match: & bear in wave \Bstrut\\
			\cline{2-9} 
			& \includegraphics[width=\figwidthvis, height=1.8cm, align=c]{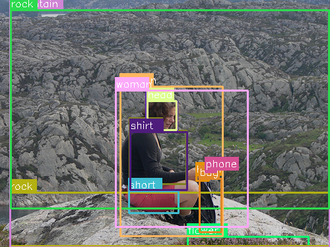} &
			\includegraphics[width=\figwidthvis, align=c]{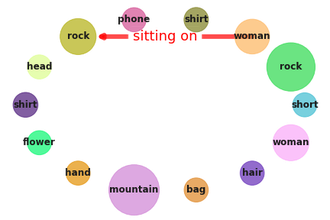} & & \includegraphics[width=\figwidthvis, height=1.8cm, align=c]{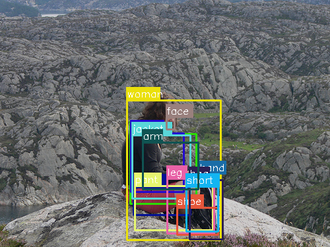} &
			\includegraphics[width=\figwidthvis, align=c]{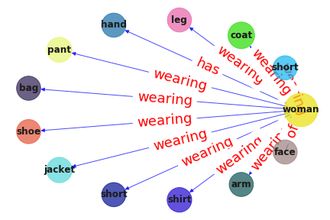} & 
			& 
			\includegraphics[width=\figwidthvis, height=1.8cm, align=c]{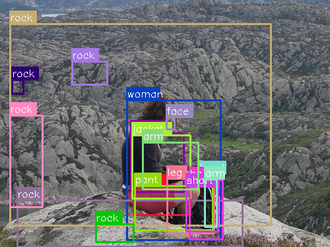} & 
			\includegraphics[width=\figwidthvis, align=c]{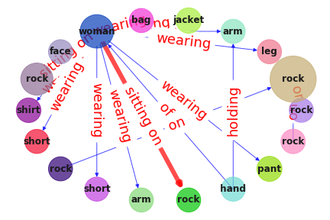} \Tstrutmore\Bstrut\\
			& zero-shot triplet: & {woman sitting on rock} & & \multicolumn{2}{r|}{no triplet involving a rock in top-20} & & \multicolumn{2}{r}{match: woman sitting on rock}\\
			\hline
		\end{tabular}
		\vspace{-10pt}
		\caption{\small Visualizations of scene graph generation for zero-shots (denoted as thick red arrows) on Visual Genome using Message Passing with the baseline loss versus our loss. We show two cases: (\textbf{top four rows}) when the baseline model makes a correct prediction, while our model is incorrect; and (\textbf{bottom four rows}) when the baseline is incorrect, while ours is correct. For the purpose of this visualization, a prediction is considered correct when the zero-shot triplet is in the top-20 triplets in the image regardless if the detected bounding boxes overlap with the ground truth. In the first case (when the baseline is correct), the predicted triplet often includes the `on' predicate, which we believe is due to the baseline being more biased to the frequency distribution. The model with our loss makes more diverse predictions showing a better understanding of scenes. Also, the ground truth is often mislabeled (see `leaf on bike') or a synonym is predicted by our model (\eg~a plant and a flower), which counts as an error. In the second case (when ours is correct), the baseline model tends to return a poor ranking of triplets and often simply gives a higher rank to frequent triplets. Conventionally~\cite{xu2017scene,zellers2018neural}, triplets are ranked according to the product of softmax scores of the subject, object and predicate. Most edges are two-way, but for clarity we show them one-way.
			{These examples are picked randomly. Intended to be viewed on a computer display.}}
		\label{fig:example}
	\end{figure}
	
	\section{Conclusions}
	Scene graphs are a useful semantic representation of images, accelerating research in many applications including visual question answering.
	It is vital for the SGG model to perform well on unseen or rare compositions of objects and predicates, which are inevitable due to an extremely long tail of the distribution over triplets. We show that strong baseline models do not effectively learn from all labels, leading to poor generalization on few/zero shots. Moreover, current evaluation metrics do not reflect this problem, exacerbating it instead.
	We also show that learning well from larger graphs is essential to enable stronger generalization. To this end, we modify the loss commonly used in SGG and achieve significant improvements and, in certain cases, state-of-the-art results, on both the existing and our novel weighted metric.
	
	\section*{\small Acknowledgments}
	BK is funded by the Mila internship, the Vector Institute and the University of Guelph. CC is funded by DREAM CDT. EB is funded by IVADO. This research was developed with funding from DARPA. The views, opinions and/or findings expressed are those of the authors and should not be interpreted as representing the official views or policies of the Department of Defense or the U.S.~Government. The authors also acknowledge support from the Canadian Institute for Advanced Research and the Canada Foundation for Innovation. We are also thankful to Brendan Duke for the help with setting up the compute environment. Resources used in preparing this research were provided, in part, by the Province of Ontario, the Government of Canada through CIFAR, and companies sponsoring the Vector Institute: \url{http://www.vectorinstitute.ai/#partners}.
	\medskip
	\small
	
	\bibliographystyle{abbrvnat}
	\bibliography{ref}
	
	\ifarxiv 
	
	\newpage
	
	\newif\ifarxiv
\arxivtrue

\ifarxiv
\section{Appendix}
\else 

\documentclass{bmvc2k}

\begin{document}
	
	\maketitle
	\setcounter{section}{1}

	\fi
	
	\subsection{Additional Results and Analysis}

	\begin{figure}[h!]
		\newcommand{\figwidth}{3cm}
		\newcommand\Tstrutmore{\rule{0pt}{10ex}}
		\centering
		\small
		\setlength{\tabcolsep}{0pt}
		\begin{tabular}{cc|p{0.5cm}cc}
			\includegraphics[width=\figwidth,align=c]{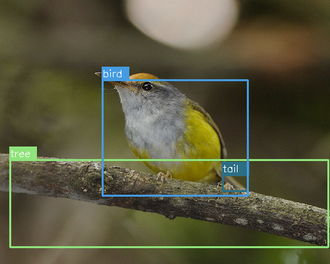} &
			\includegraphics[width=\figwidth,align=c]{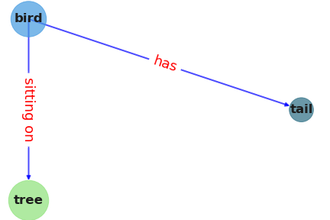} & &
			\includegraphics[width=\figwidth,align=c]{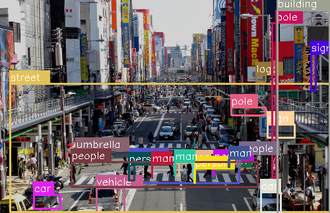} & 
			\includegraphics[width=\figwidth,align=c]{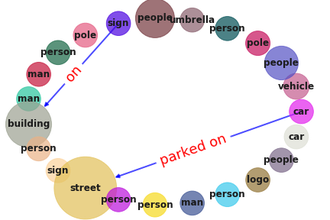} \Bstrut\\
			\includegraphics[width=\figwidth,align=c]{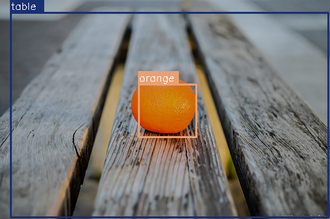} &
			\includegraphics[width=\figwidth,align=c]{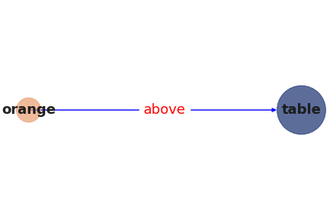} & & 
			\includegraphics[width=\figwidth,align=c]{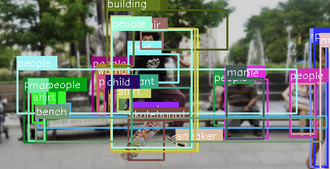} & 
			\includegraphics[width=\figwidth,align=c]{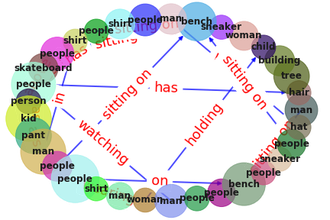} \Tstrutmore\Bstrut\\
			\includegraphics[width=\figwidth,align=c]{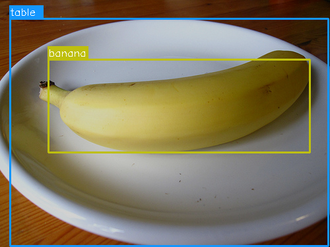} &
			\includegraphics[width=\figwidth,align=c]{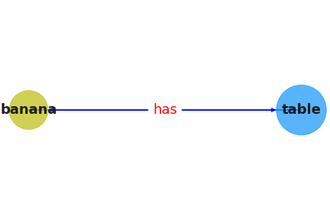} & & 
			\includegraphics[width=\figwidth,align=c]{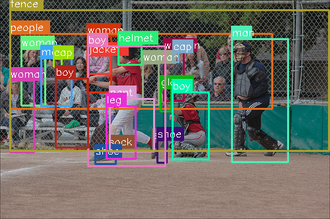} & 
			\includegraphics[width=\figwidth,align=c]{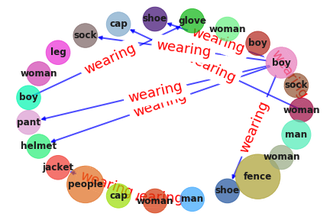} \Tstrutmore\Bstrut\\
			\hline
			\includegraphics[width=\figwidth,align=c]{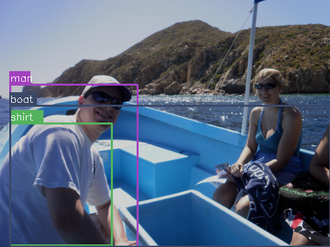} &
			\includegraphics[width=\figwidth,align=c]{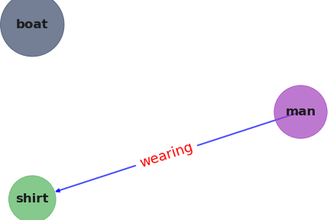} & & 
			\includegraphics[width=\figwidth,align=c]{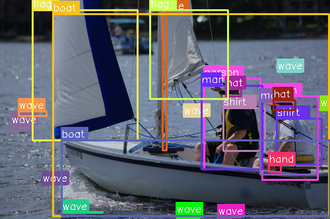} & 
			\includegraphics[width=\figwidth,align=c]{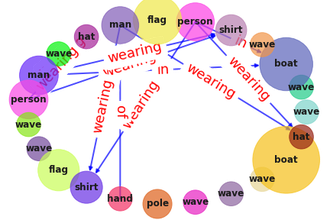} \Tstrutmore\Bstrut\\
			\hline 
			(a) & (b) & & (c) & (d) \Tstrut \Bstrut \\
		\end{tabular}
		\caption{\small Examples of small (b) and large (d) training scene graphs and corresponding images ((a) and (c) respectively) from Visual Genome (split~\cite{xu2017scene}). Two related factors mainly affect the size of the graphs: 1) the complexity of a scene (compare left and right images in the top three rows); and 2) the amount of annotations (\eg~in the bottom row two images have similar complexity, but in one case the annotations are much more sparse).
			On average, for more complex scenes, the graphs tend to be larger and more sparse ($d\leq2\%$ on the depicted large graphs, (d)), because for many reasons it is challenging to annotate all edges.
			In contrast, simple scenes can be described well by a few nodes, making it easier to label most of the edges, which makes graphs more dense ($d>15\%$ on the depicted small graphs, (b)).\looseness-1}
		\label{fig:small_large_graphs}
		\vspace{-10pt}
	\end{figure}

	\begin{figure}[h!]
		\newcommand{\figwidth}{3cm}
		\newcommand\Tstrutmore{\rule{0pt}{10ex}}
		\centering
		\small
		\setlength{\tabcolsep}{0pt}
		\begin{tabular}{cc|p{0.5cm}cc}
			\includegraphics[width=\figwidth,align=c]{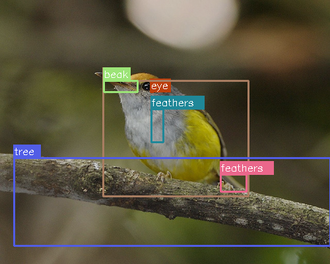} &
			\includegraphics[width=\figwidth,align=c]{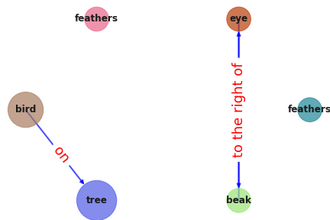} & & 
			\includegraphics[width=\figwidth,align=c]{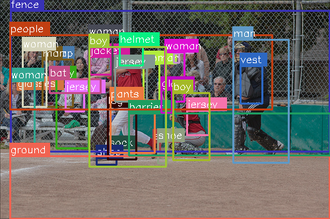} & 
			\includegraphics[width=\figwidth,align=c]{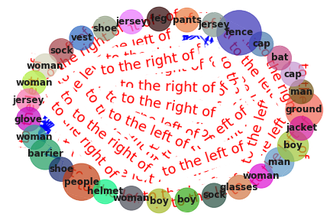} \Bstrut\\
			\includegraphics[width=\figwidth,align=c]{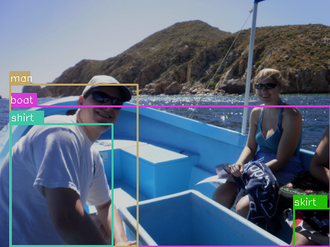} &
			\includegraphics[width=\figwidth,align=c]{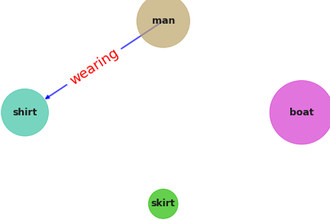} & & 
			\includegraphics[width=\figwidth,align=c]{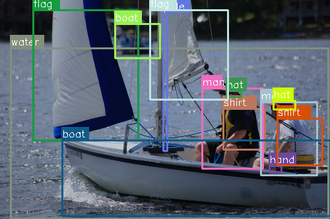} & 
			\includegraphics[width=\figwidth,align=c]{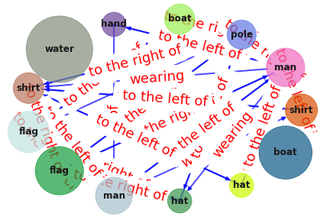} \Tstrutmore \Bstrut\\
			\hline 
			(a) & (b) & & (c) & (d) \Tstrut \\
		\end{tabular}
		\caption{\small For comparison, we illustrate scene graphs of the same images, but annotated in GQA. As opposed to VG, scene graphs in GQA are generally larger and more dense, primarily due to ``left of'' and ``right of'' edge annotations (see detailed dataset statistics in Table~\ref{table:datasets}).}
		\label{fig:small_large_graphs_gqa}
		\vspace{-10pt}
	\end{figure}

	\begin{figure}
		\centering
		\begin{small}
			\setlength{\tabcolsep}{5pt}
			\vspace{-5pt}
			\begin{tabular}{ccc}
				\scriptsize $b = 1$ & \scriptsize $b = 6$ & \scriptsize $b = 6$, edge sampling \\
				\hline 
				\includegraphics[height=2.7cm,align=c]{VG_graphs_bg_b1} &
				\includegraphics[height=2.5cm,align=c]{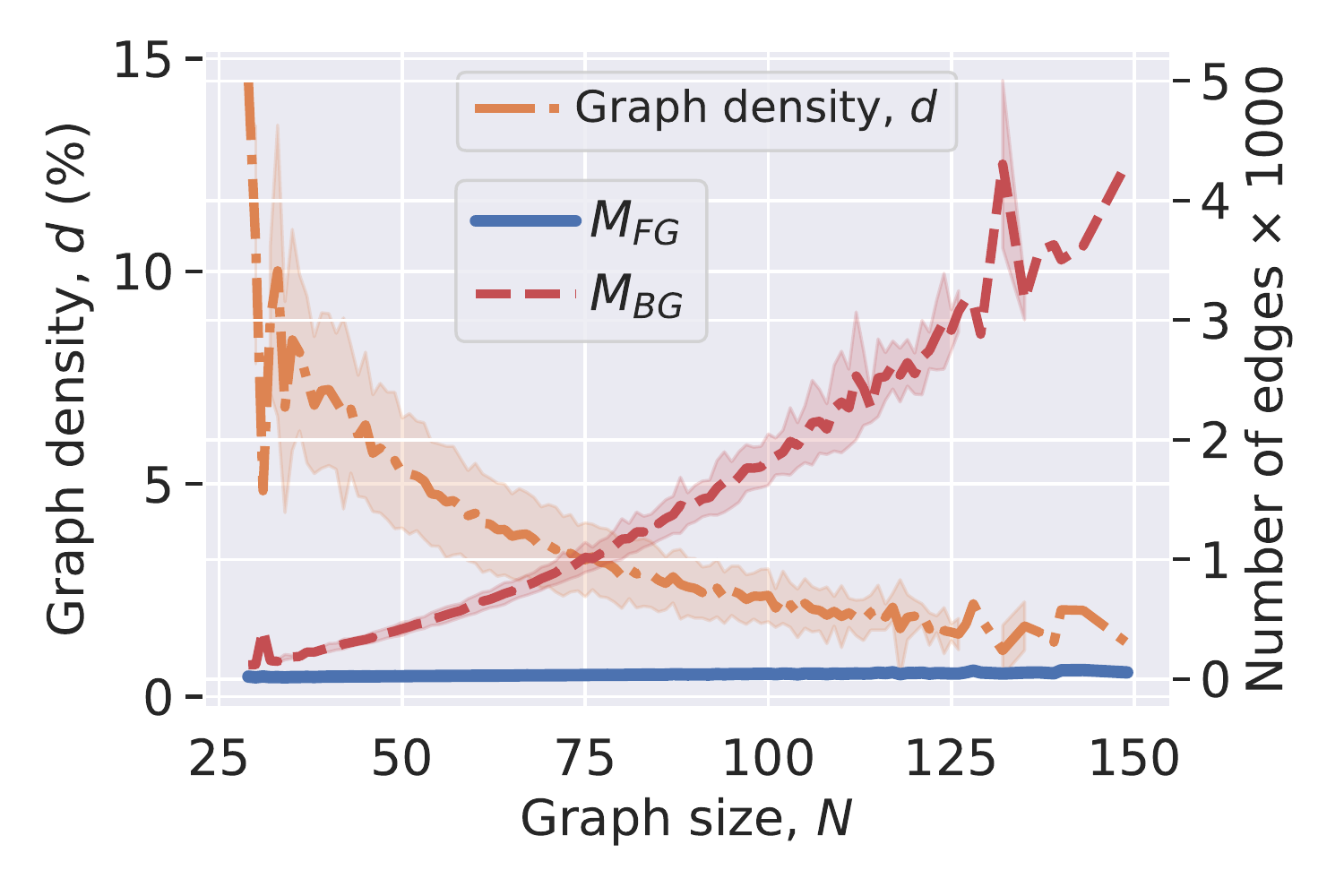} &
				\includegraphics[height=2.5cm,align=c]{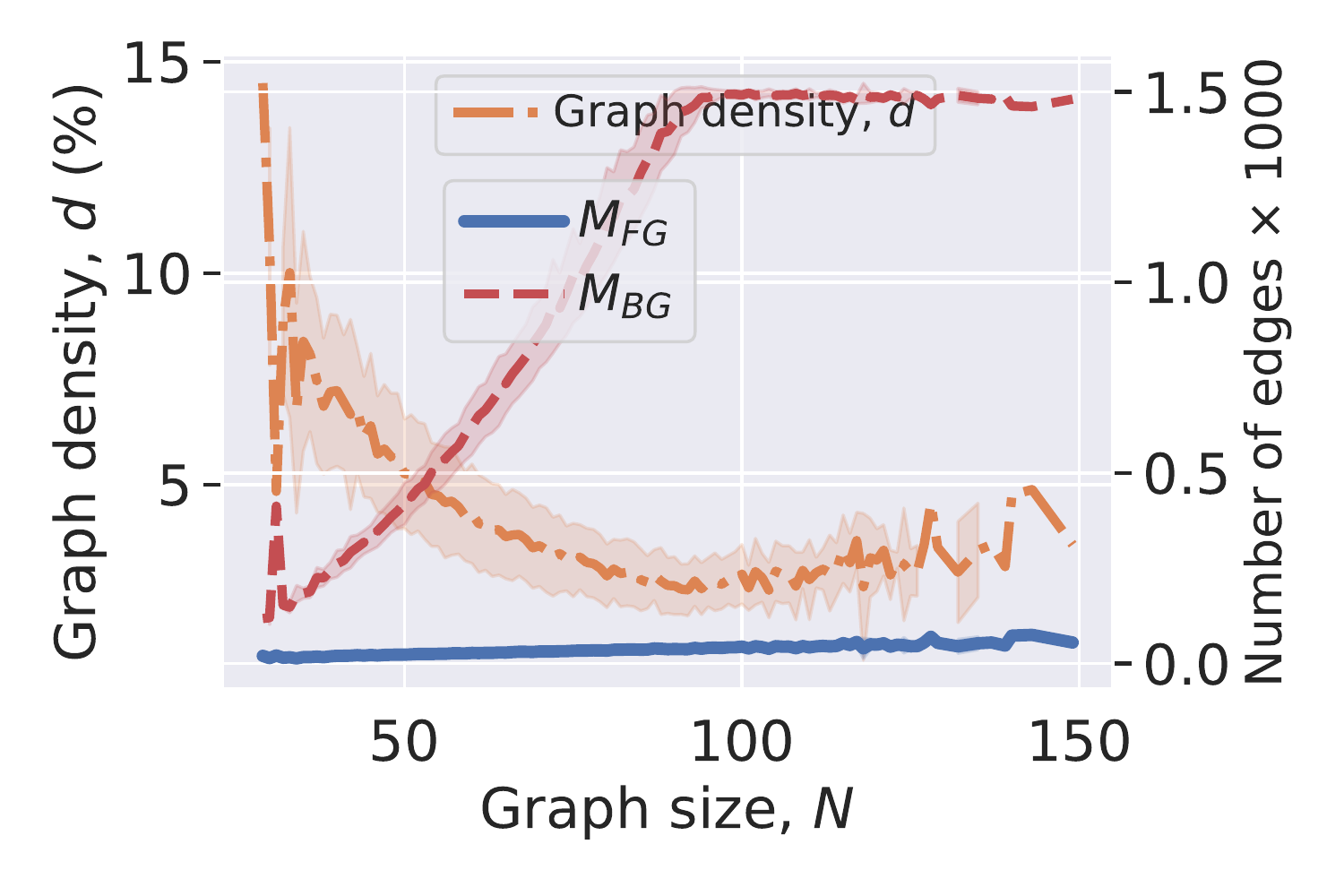} \\ 
				\includegraphics[height=2.5cm,align=c]{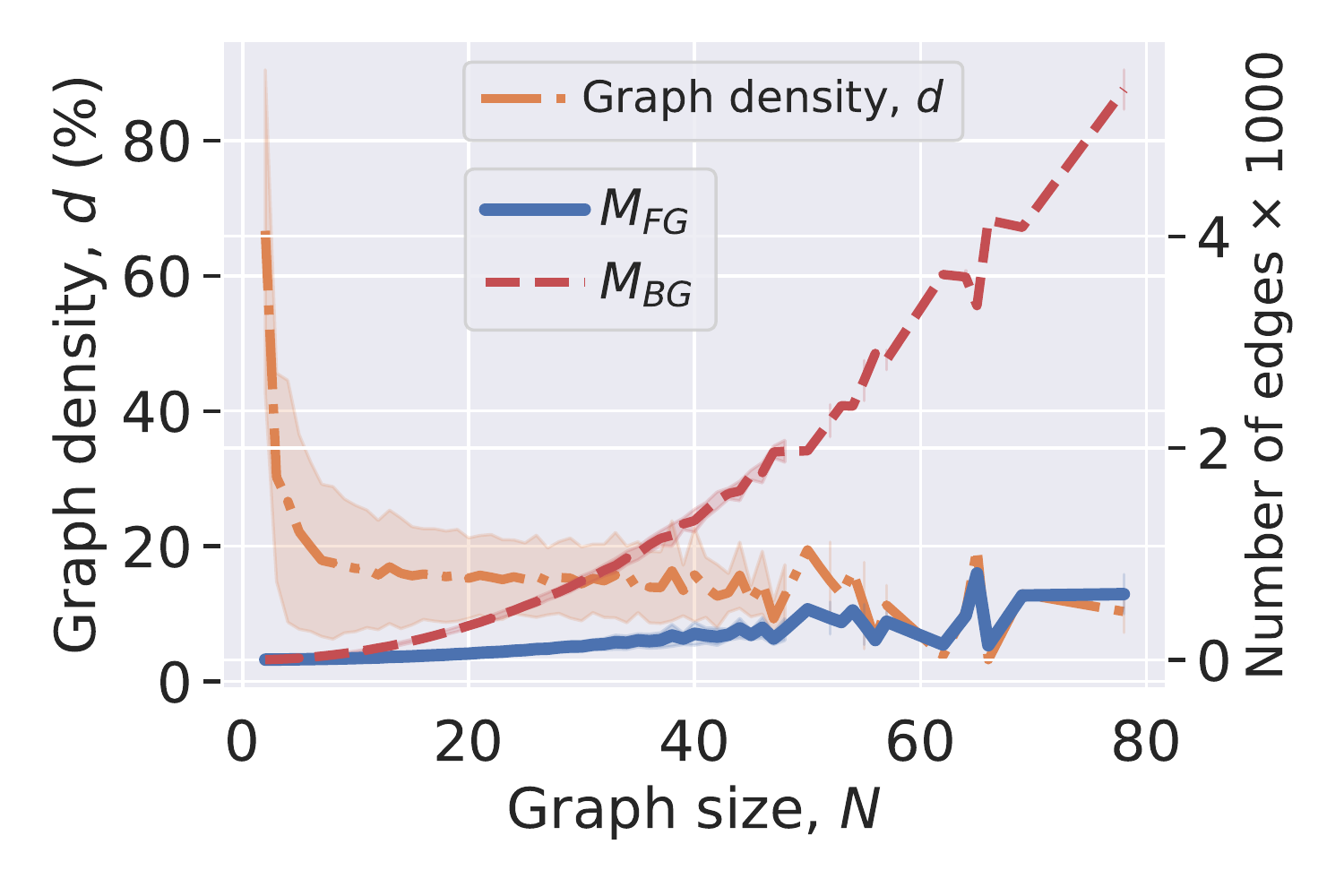}
				& \includegraphics[height=2.5cm,align=c]{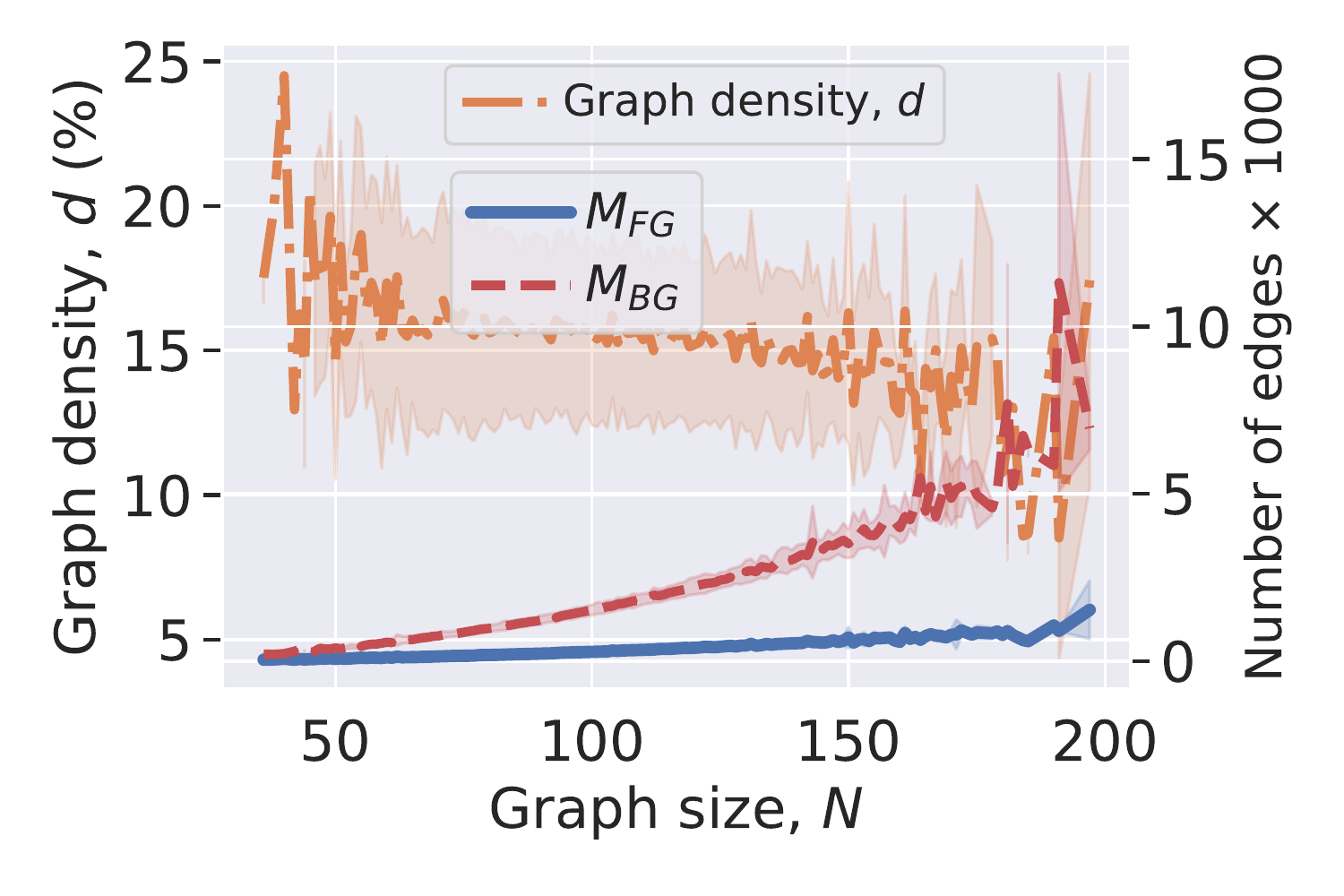} & \includegraphics[height=2.5cm,align=c]{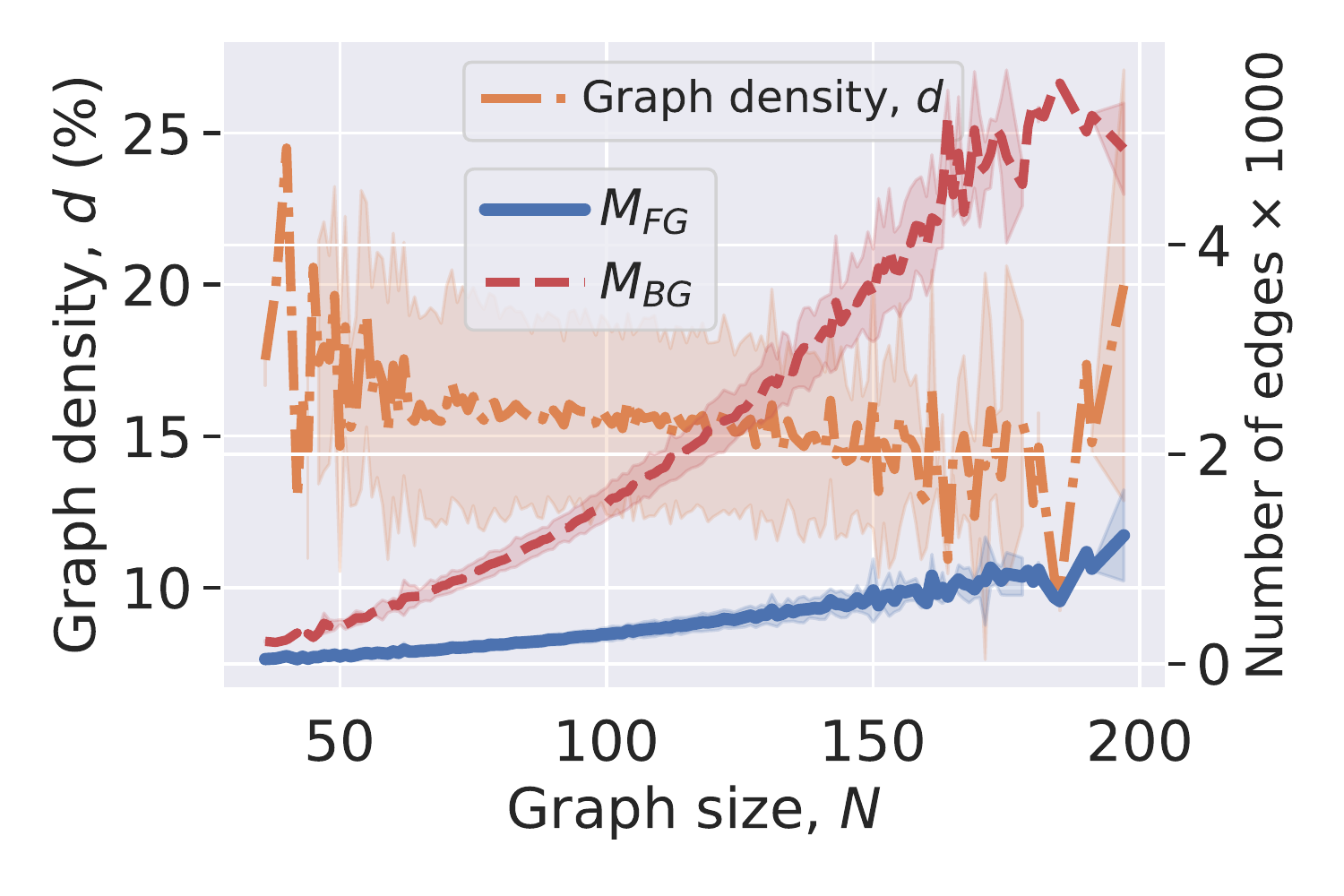} 
			\end{tabular}
		\end{small}
		\vspace{-5pt}
		\caption{
			\small Plots additional to Figure~\ref{fig:graph_density}. Graph density $d$ of a batch of scene graphs for different batch sizes $b$ for VG (top row) and GQA (bottom row). For $b=1$: $M_{FG}\approx 0.5N$ for VG and $M_{FG}\approx 8N$. Larger batches make the density of a batch less variable and, on average, larger. This is because by stacking scene graphs in a batch, there are no BG edges between different graphs, so $M_{BG}$ grows slower making the density larger. Thus, increasing the batch size partially fixes the discrepancy of edge losses between small and large scene graphs. Subsampling edges during training also helps to stabilize and increase graph density. We explore this effect in more detail on Figure~\ref{fig:results_graph_size_more}.
		}
		\vspace{-5pt}
		\label{fig:graph_density_more}
	\end{figure}
	
	\begin{figure}[h!]
		\centering
		\begin{tiny}
			\setlength{\tabcolsep}{5pt}
			\begin{tabular}{ccc}
				\multicolumn{3}{c}{{\includegraphics[width=0.4\textwidth,trim={14cm 8.5cm 1.5cm 0.5cm},clip]{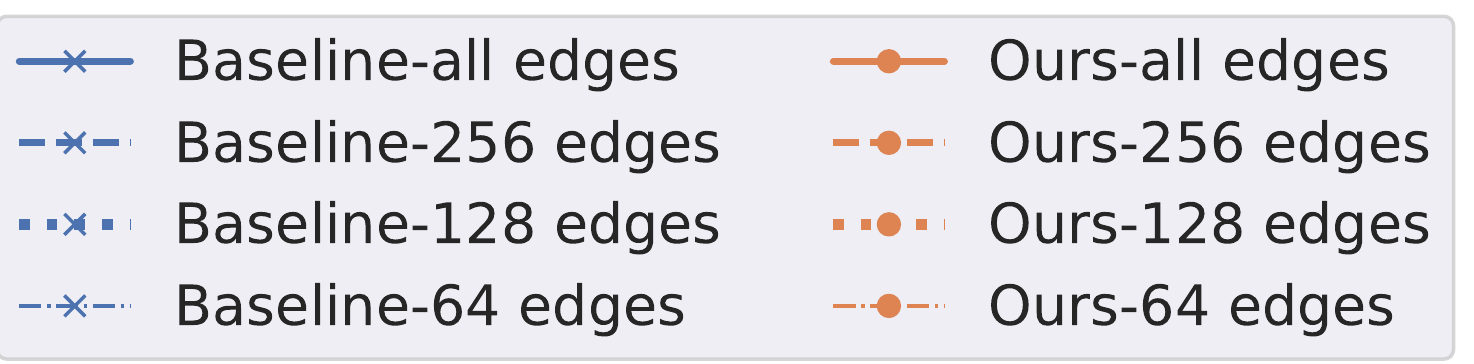}}} \\
				\includegraphics[height=2.5cm,align=c]{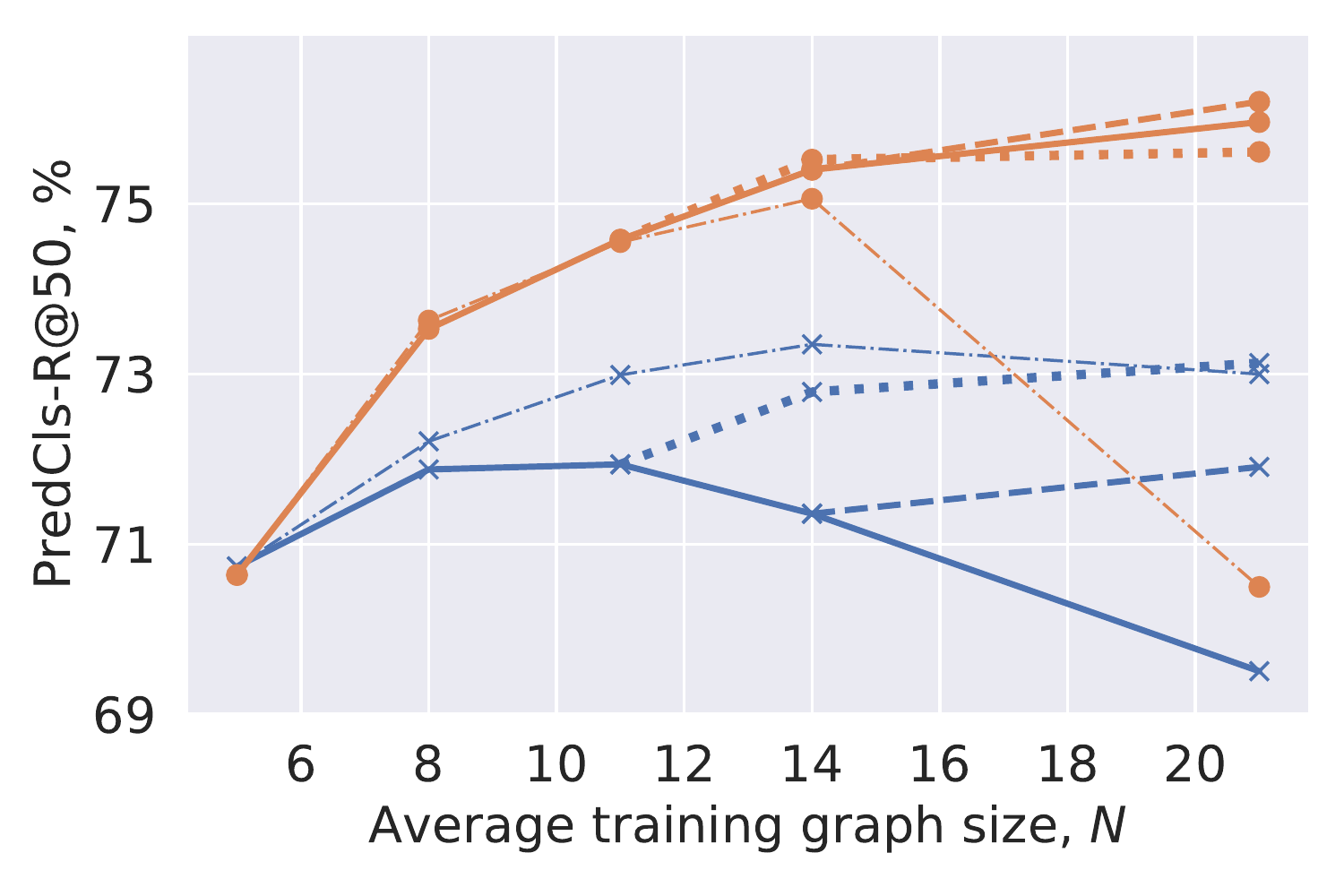} & \includegraphics[height=2.5cm,align=c]{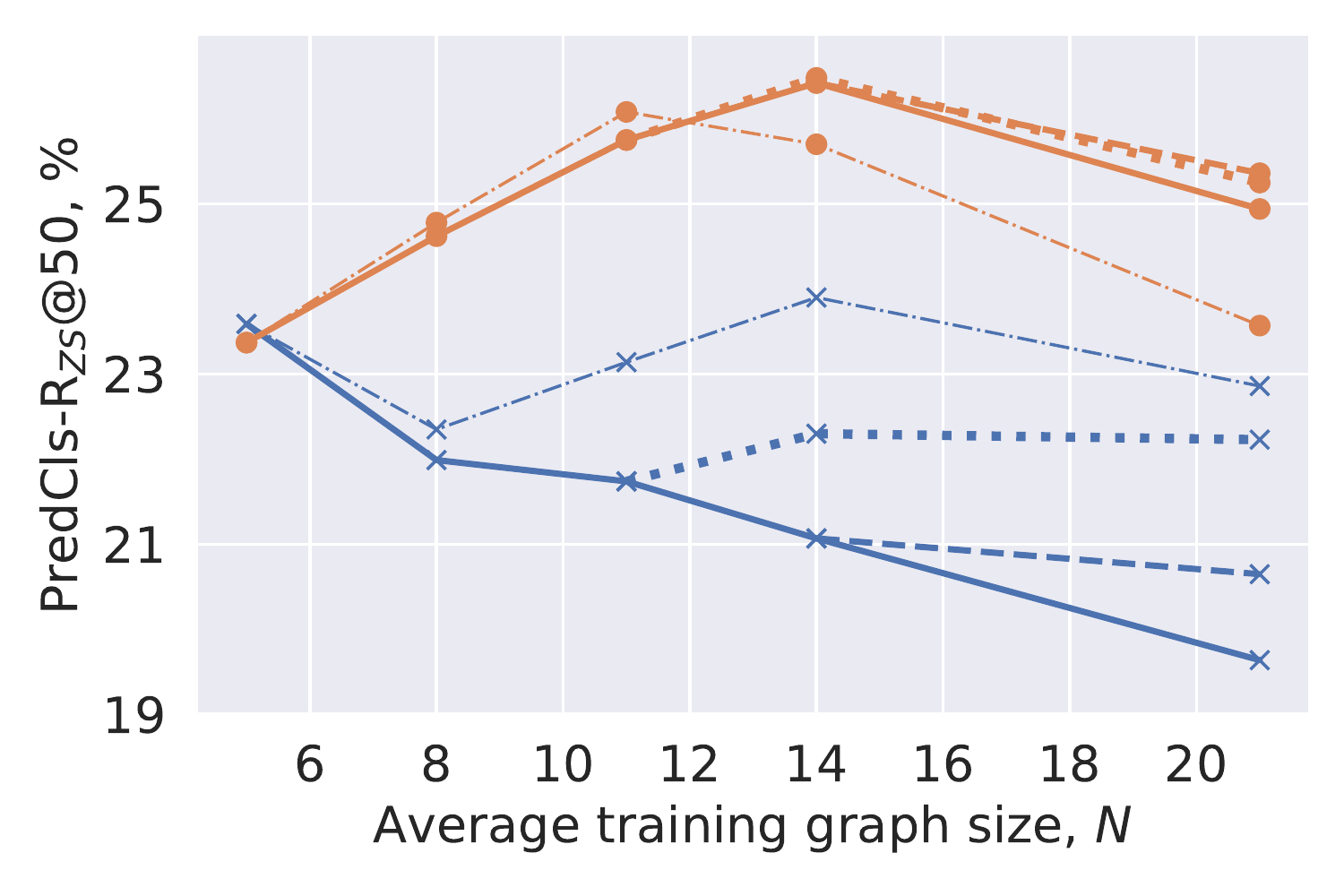} & \includegraphics[height=2.5cm,align=c]{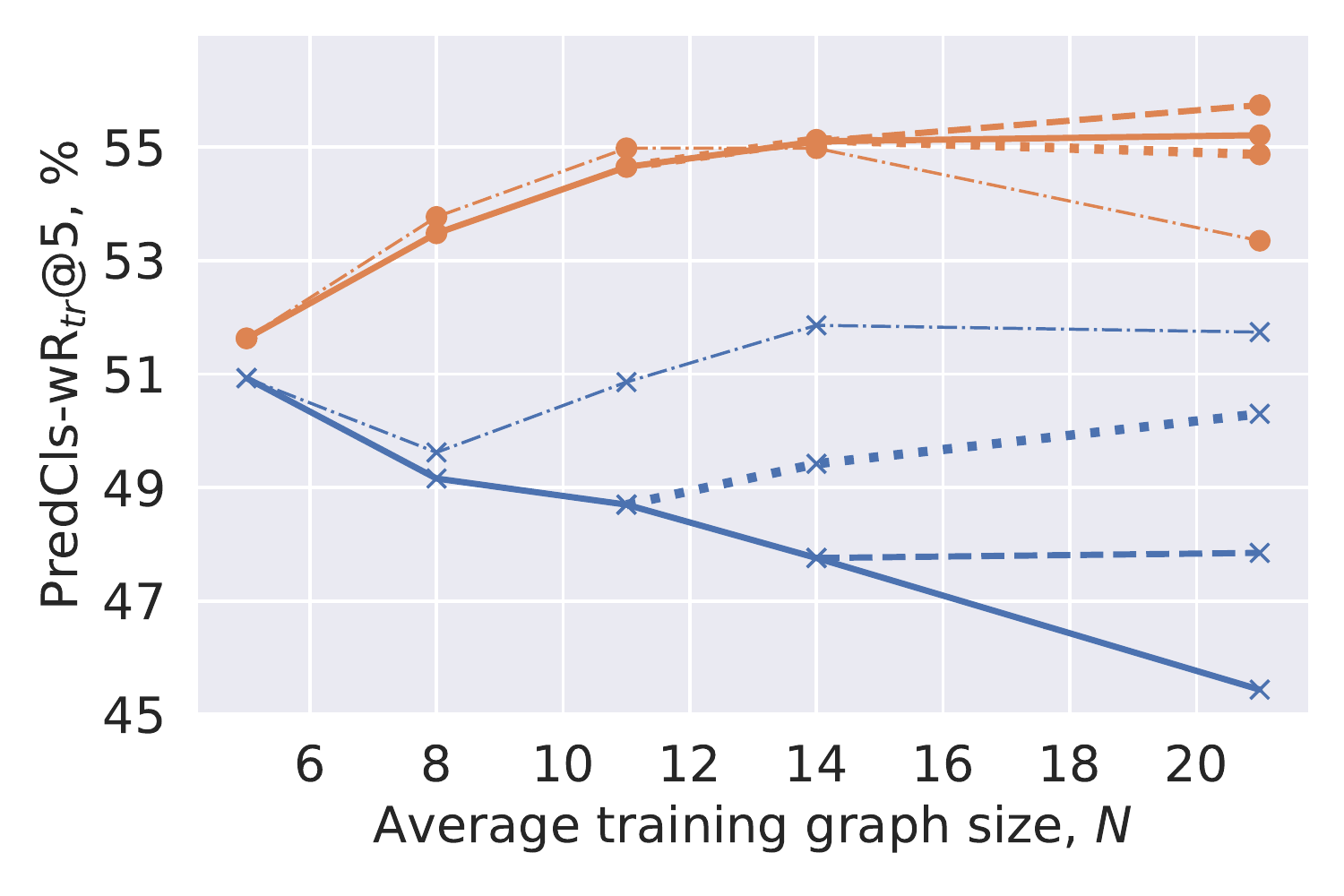} \\
				\includegraphics[height=2.5cm,align=c]{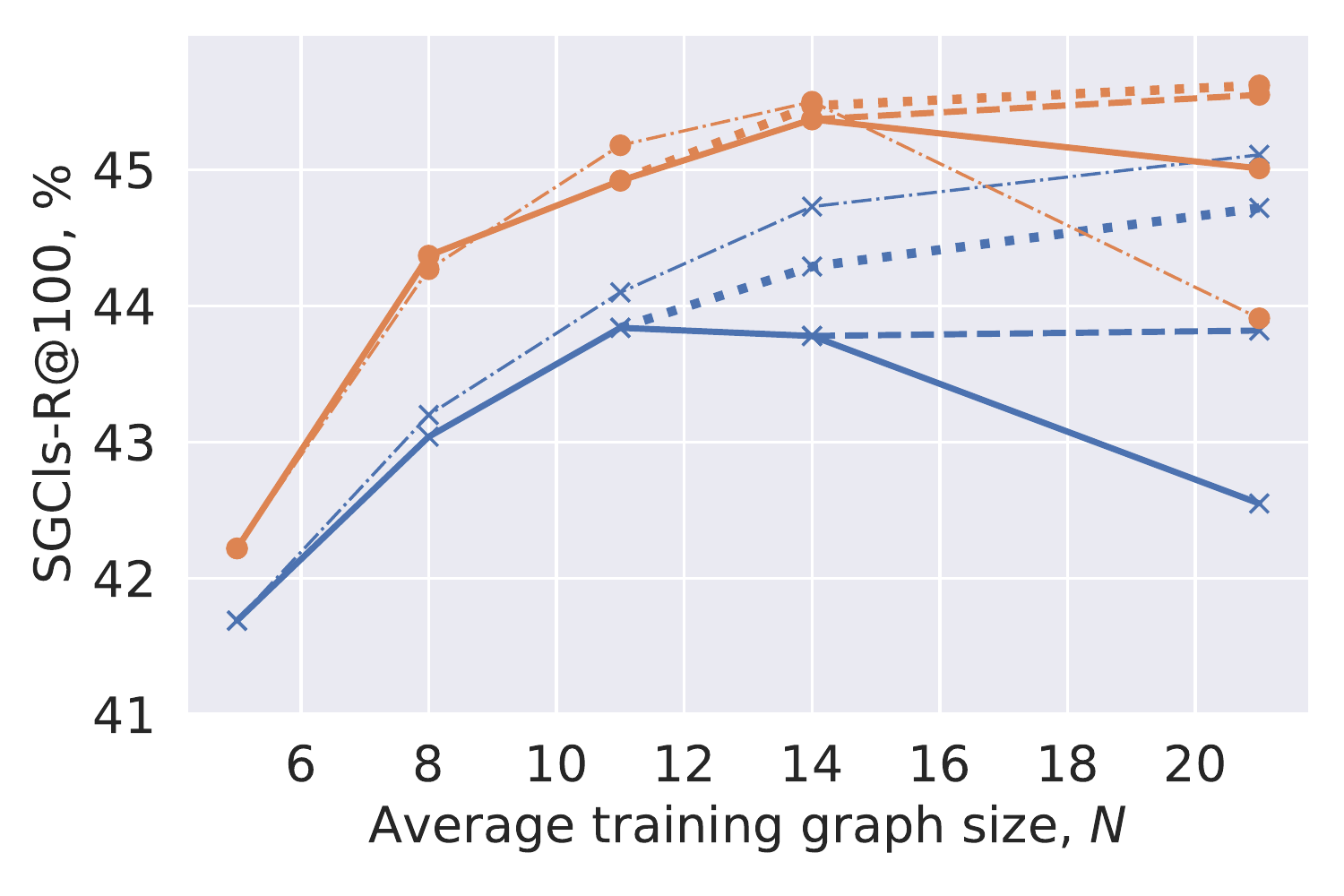} & \includegraphics[height=2.5cm,align=c]{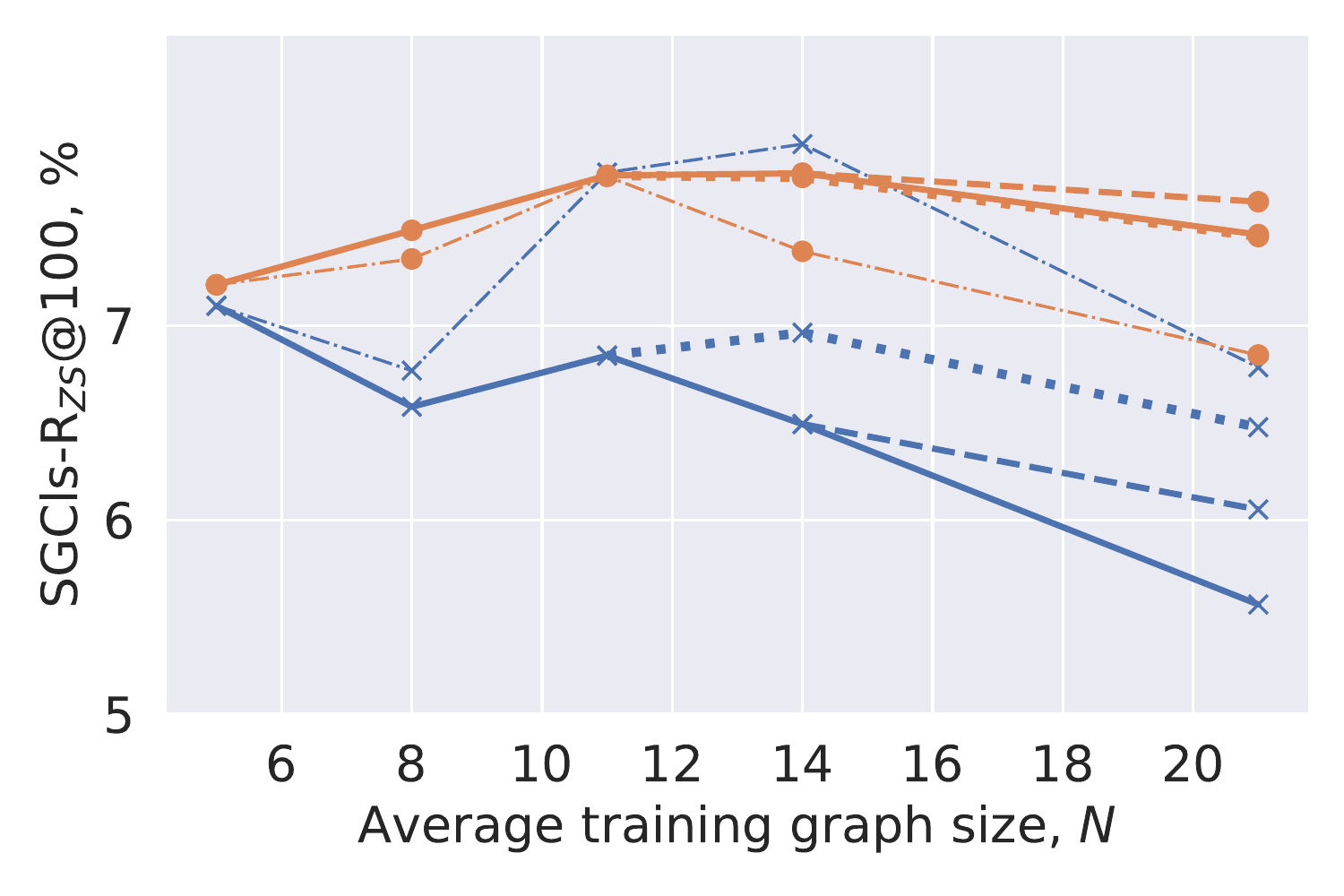} & \includegraphics[height=2.5cm,align=c]{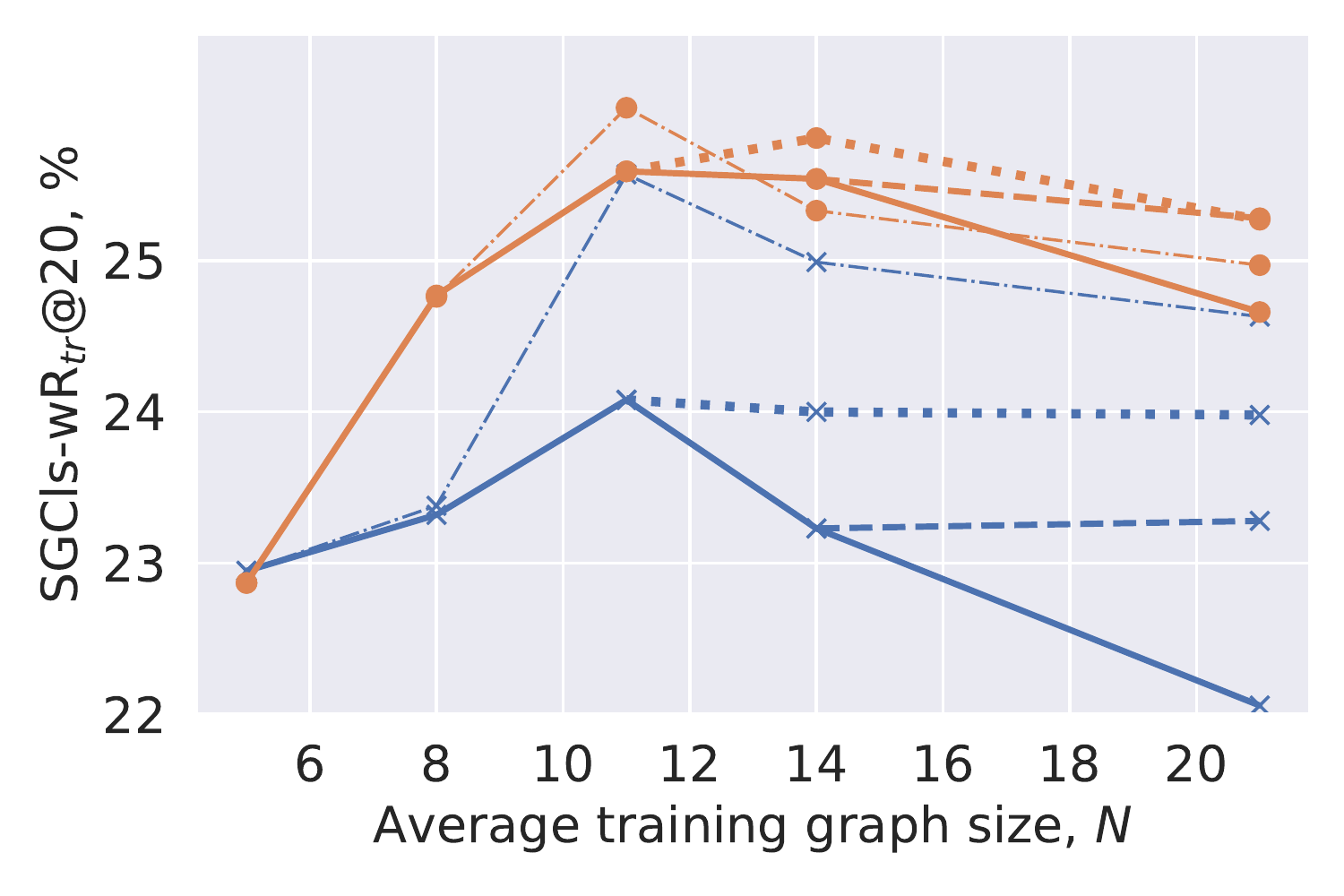} \\
			\end{tabular}
		\end{tiny}
		\caption{\small 
			In these experiments, we split the training set of Visual Genome (split~\cite{xu2017scene}) into five subsets with different graph sizes in each subset. This allows us to study how well the models learn from graphs of different sizes.
			\textbf{Edge sampling.} One way to stabilize and increase graph density is to sample a fixed number of FG and BG edges (on the legend, we denote the total possible number of edges per image, $M$). So we compare results with and without this sampling. Sampling improves baseline results of models trained on larger graphs, however, our best results are still consistently higher than the best baseline results. Even though sampling partially solves the problem of varying density, it does not solve it in a principle way as we do. Sampling only sets the upper bound on the number of edges, so some batches have fewer edges, creating a discrepancy between the losses of edges. Setting a lower bound on the number of edges is challenging, because some graphs are very sparse, meaning that frequently many more graphs need to be sampled to get enough edges, which requires much more computational resources. 
			\textbf{Potential oversmoothing.} In case of our loss the results for larger graphs do not improve in some cases or even slightly get worse. We believe that besides density normalization, another factor making it challenging to learn from large graphs, can be related to the ``oversmoothing'' effect~\cite{rong2019truly}. Oversmoothing occurs when all nodes after the final graph convolution start to have very similar features. This typically happens in deep graph convolutional networks. But oversmoothing can also occur in \textit{complete} graphs, which are used in the SGG pipeline as the input to message passing (Figure~\ref{fig:overview}). Complete graphs lead to node features being pooled (averaged) over a very large neighborhood (i.e. all other nodes) and averaging over too many node features is detrimental to their discriminative content. A direction for resolving this issue can be using some form of edge proposals and attention over edges~\cite{yang2018graph,velivckovic2017graph}.
		}
		\vspace{-10pt}
		\label{fig:results_graph_size_more}
	\end{figure}
	
	\newpage
	\begin{figure}[h!]
		\centering
		\begin{tiny}
			\setlength{\tabcolsep}{1pt}
			\begin{tabular}{cccc}
				\includegraphics[height=2.5cm,align=c]{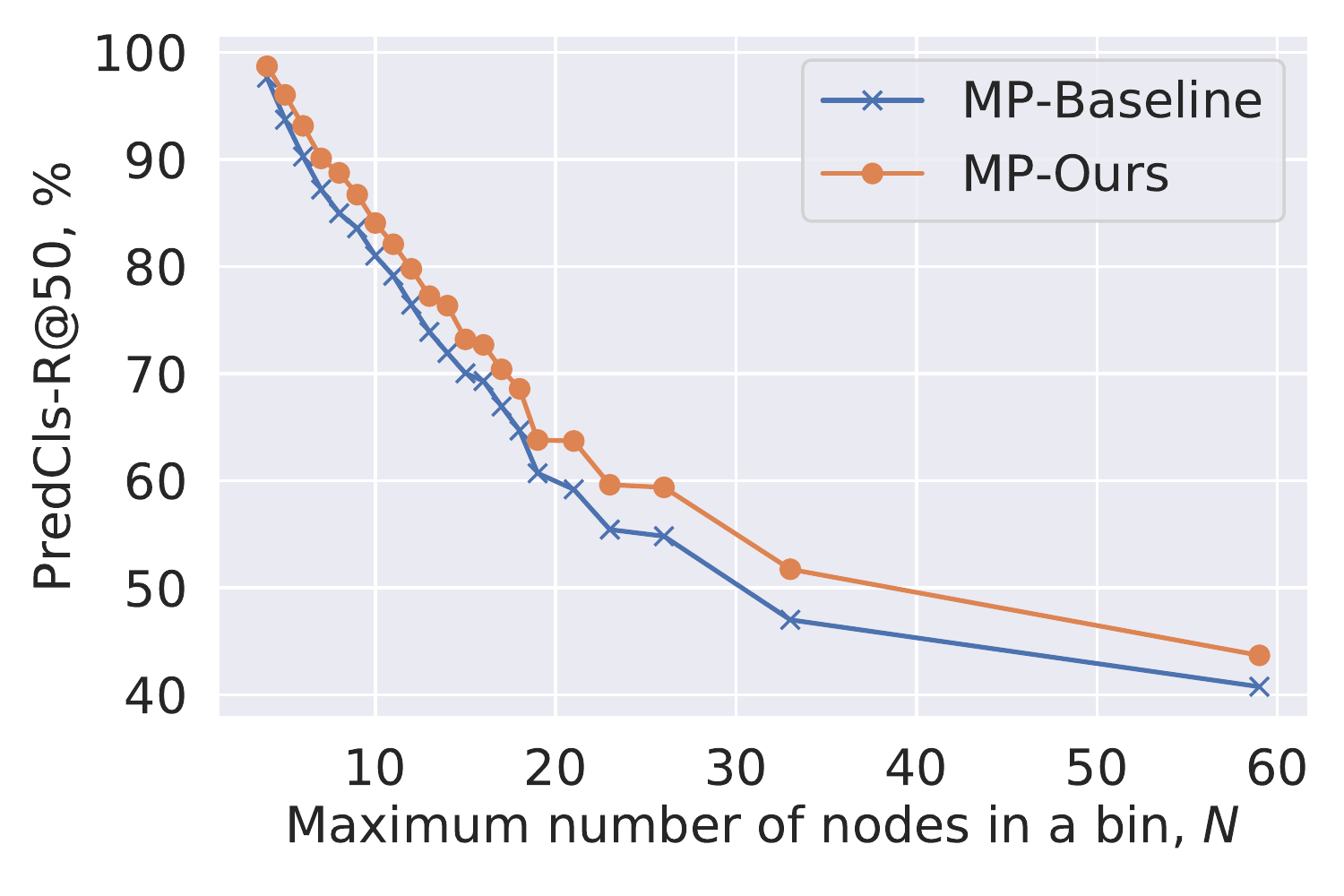} & 
				\includegraphics[height=2.5cm,align=c]{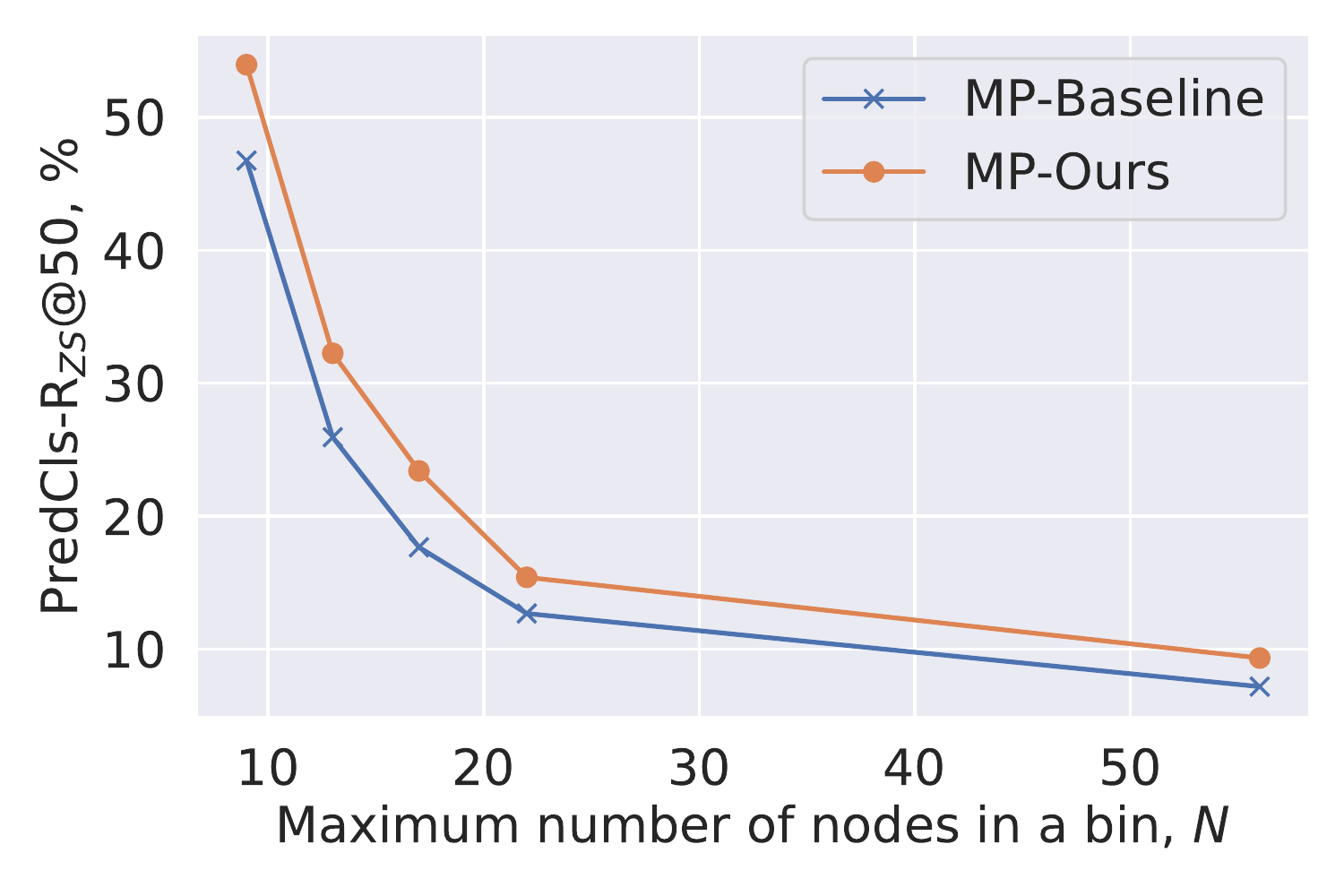} &
				\includegraphics[height=2.5cm,align=c]{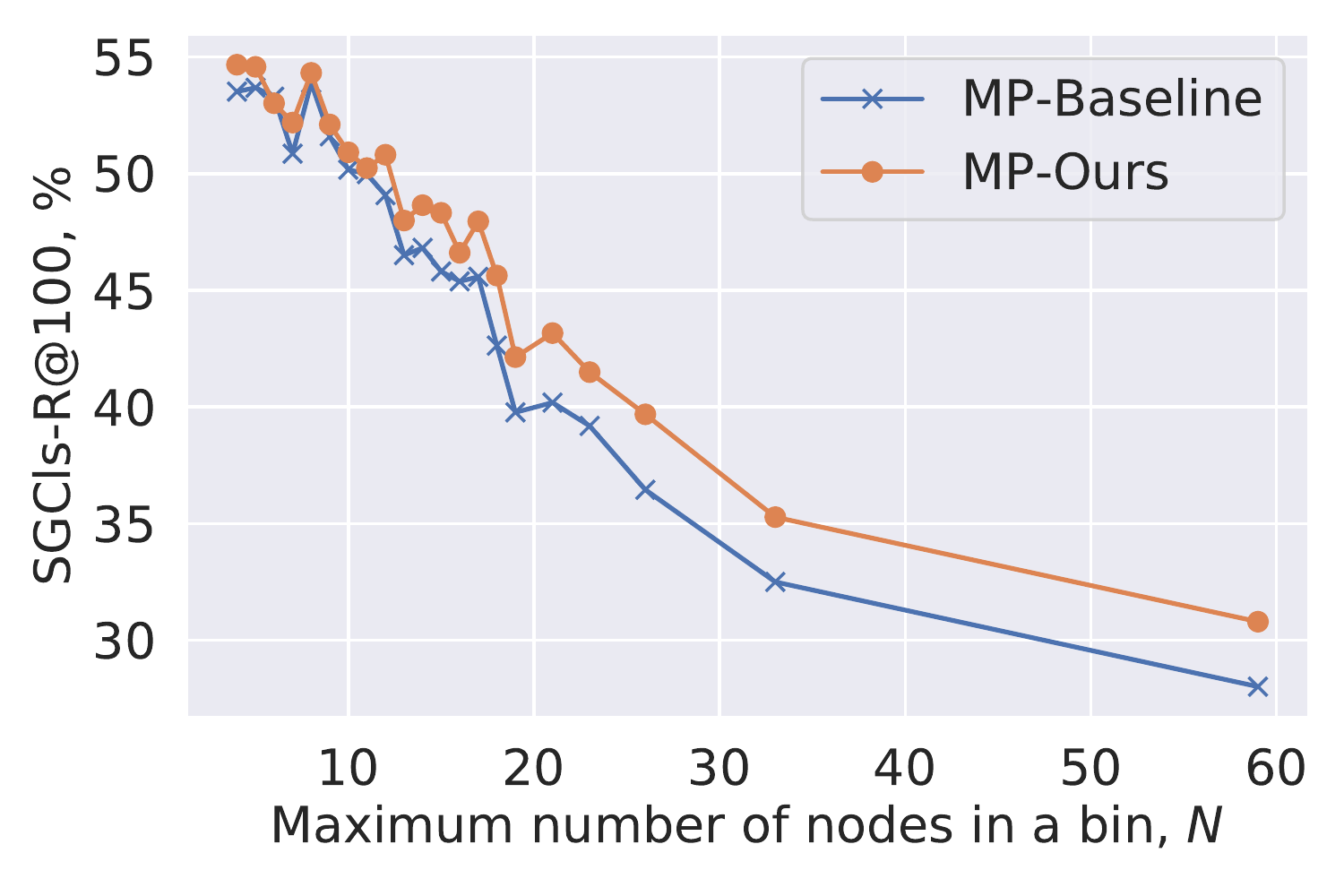} & 
				\includegraphics[height=2.5cm,align=c]{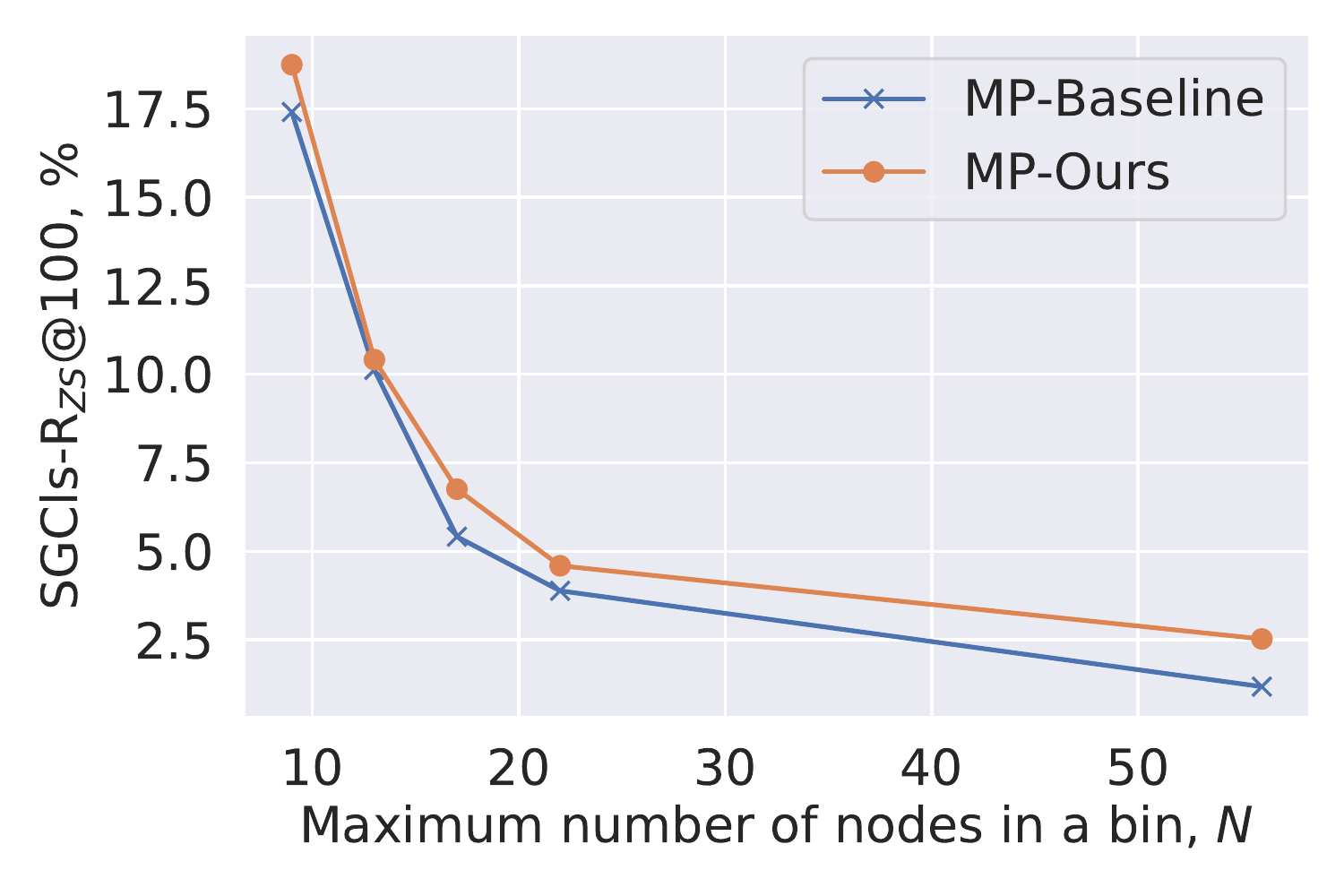}  \\
			\end{tabular}
		\end{tiny}
		\vspace{-5pt}
		\caption{\small In these experiments, we take a Message Passing model, trained on all scene graphs of Visual Genome, and test it on graphs of different sizes. To plot the curves, we sort the original test set by the number of nodes in scene graphs and then the split the sorted set into several bins with an equal number of scene graphs in each bin (~1000 per bin). Each point denotes an average recall in a bin.
			As was shown in Figure~\ref{fig:results_graph_size_more}, our loss makes learning from larger graphs more effective. Here, we also show that our loss makes the models perform better on larger graphs at test time. We believe that since our loss penalizes larger graphs more during training, the model better learns how to process large graphs. However, we observe the performance drops for larger graphs both for the baseline and our loss. One of the reasons for that might be related to oversmoothing, which can be more pronounced in large graphs (also discussed in Figure~\ref{fig:results_graph_size_more}). Another reason can be a lack of large graphs (\eg > 40 nodes) in the training set. This would align with prior work~\cite{knyazev2019understanding}, showing that generalization to larger graphs is challenging and proposed attention as a way for addressing it. This problem, in the context of scene graphs, can be addressed in future work. For example, one interesting study may include training on so called ``region graphs'', describing small image regions, available in Visual Genome~\cite{Krishna_2017} and attempting to predict full scene graphs.}
		\label{fig:results_graph_size_test}
		\vspace{-10pt}
	\end{figure}

	\begin{table}[h!] %
		\vspace{-5pt}
		\begin{scriptsize}
			\setlength{\tabcolsep}{8pt}
			\begin{tabular}{p{4cm}p{2.4cm}p{2.4cm}p{2.4cm}p{2.4cm}}
				& \textbf{VG}~\cite{xu2017scene} & \textbf{VTE}~\cite{zhang2017visual} & \textbf{GQA}~\cite{hudson2019gqa} & \textbf{GQA-nLR} \\
				\hline \\
				\#obj classes & 150 & 200 & 1,703 & 1,703 \\
				\#rel classes & 50 & 100 & 310 & 308 \\
				\# train images & 57,723 & 68,786 & 66,078 & 59,790  \\
				\# train triplets (unique) & 29,283 & 19,811 & 470,129 & 98,367 \\
				\# val images & 5,000 & 4,990 & 4,903 & 4,382 \\
				\# test images & 26,446 & 25,851 & 10,055 & 9,159 \\
				\# test-ZS images & 4,519 & 653 & 6,418 & 4,266 \\
				\# test-ZS triplets (unique/total) & 5,278/7,601 & 601/2,414 & 37,116/45,135 & 9,704/11,067 \\
				\hline \\
			\end{tabular}
			\newline
			\begin{tabular}{p{4cm}llllllll}
				& \tiny min-max & \tiny avg\std{std} & \tiny min-max & \tiny avg\std{std} & \tiny min-max & \tiny avg\std{std} & \tiny min-max & \tiny avg\std{std} \\
				\cline{2-9} \\
				$N$ train & 2-62 & 12\std{6} & 2-98 & 13\std{9} & 2-126 & 17\std{8} & 2-126 & 17\std{8} \\
				$d$ (\%) train & 0.04-100  & 7\std{8} & 0.7-100  & 12\std{13} & 0.5-100 & 17\std{10} & 0.03-100 & 3\std{5} \\
				\hline
				$N$ test & 2-58  & 12\std{7} & 2-110  & 13\std{9} & 2-97 & 17\std{8} & 2-67 & 17\std{8}  \\
				$d$ (\%) test & 0.12-100 & 6\std{8} & 0.6-100 & 11\std{12} & 0.6-100 & 17\std{10} & 0.05-100 & 3\std{5} \\
				\hline
				$N$ test-ZS & 2-55 & 14\std{7} & 2-78 & 8\std{11} & 2-97 & 19\std{8} & 2-65 & 18\std{8} \\
				$d$ (\%) test-ZS & 0.05-50 & 2\std{4} & 0.05-100 & 3\std{7} & 0.03-50 & 3\std{3.6} & 0.02-50 & 1.5\std{3} \\
			\end{tabular}
		\end{scriptsize}
		\vspace{-5pt}
		\caption{\small Statistics of Visual Genome~\cite{Krishna_2017} variants used in this work. GQA-nLR is our modification of GQA~\cite{hudson2019gqa} with predicates `to the left of' and `to the right of' excluded from all training, validation and test scene graphs. Graph density $d$ decreases dramatically in this case, since those two predicates are the majority of all predicate labels.}\label{table:datasets}
	\end{table}
	
	\begin{table}[h!] %
		\vspace{-10pt}
		\begin{scriptsize}
			\setlength{\tabcolsep}{8pt}
			\begin{tabular}{p{4cm}p{2.4cm}p{2.4cm}p{2.4cm}p{2.4cm}}
				& \textbf{VG}~\cite{xu2017scene} & \textbf{VTE}~\cite{zhang2017visual} & \textbf{GQA}~\cite{hudson2019gqa} & \textbf{GQA-nLR} \\
				\hline \\
				Object detector & Faster R-CNN~\cite{ren2015faster} & \multicolumn{3}{p{7.2cm}}{Mask R-CNN~\cite{he2017mask}, chosen in lieu of Faster R-CNN, since it achieves better performance due to multitask training on COCO. In SGGen, we extract up to 50 bounding boxes with a confidence threshold of 0.2 as in~\cite{NSM2019}.} \\ \\
				Detector's backbone & VGG16 & \multicolumn{3}{c}{ResNet-50-FPN} \\ \\
				Detector pretrained on & VG~\cite{xu2017scene} & \multicolumn{3}{p{7.2cm}}{COCO (followed by fine-tuning on GQA in case of SGGen)} \\ \\
				Learning rate & $0.001 \times b$ & $0.001 \times b$ & $0.002 \times b$ (increased due to larger graphs in a batch) & $0.001 \times b$ \\ \\
				Batch size (\# scene graphs), $b$ & \multicolumn{4}{c}{6} \\ \\
				\# epochs & \multicolumn{4}{c}{MP: 20, lr decay by 0.1 after 15 epochs; NM: 12, lr decay by 0.1 after 10 epochs}
			\end{tabular}
		\end{scriptsize}
		\vspace{-5pt}
		\caption{\small Architecture details. In NM's implementation, the number of epochs is determined automatically based on the validation results. We found it challenging to choose a single metric to determine the number of epochs, so we fix the number of epochs based on manual inspection of different validation metrics.}\label{table:arch}
		\vspace{-15pt}
	\end{table}
	
	\subsection{Evaluation}
	
	\textbf{Evaluation of zero/few shot cases.} To evaluate $n$-shots using image-level recall, we need to keep in the test images only those triplets that have occurred no more than $n$ times and remove images without such triplets. This results in computing recall for very sparse annotations, so the image-level metric can be noisy and create discrepancies between simple images with a few triplets and complex images with hundreds of triplets. For example, for an image with only two ground truth triplets, R@100 of 50\% can be a quite bad result, while for an image with hundreds of triplets, this can be an excellent result. Our Weighted Triplet Recall is computed for all test triplets joined into a single set, so it resolves this discrepancy.
	
	\textbf{Constrained vs unconstrained metrics.}
	In the graph constrained case~\cite{xu2017scene}, only the top-1 predicted predicate is considered when triplets are ranked, and follow-up works~\cite{newell2017pixels,zellers2018neural} improved results by removing this constraint. This unconstrained metric more reliably evaluates models, since it does not require a perfect triplet match to be the top-1 prediction, which is an unreasonable expectation given plenty of synonyms and mislabeled annotations in scene graph datasets. For example, `man \textit{wearing} shirt' and `man \textit{in} shirt' are similar predictions, however, only the unconstrained metric allows for both to be included in ranking. The \textsc{SGDet+} metric~\cite{yang2018graph} has a similar motivation as removing the graph constraint, but it does not address the other issues of image-level metrics.
	
	\ifarxiv 
	
	\else
	
	\medskip
	\small
	
	\newpage
	\bibliography{egbib}
	
\end{document}

\fi
	
	\else
	\fi
	
\end{document}